\documentclass[lettersize,journal]{IEEEtran}

\makeatletter
\@ifundefined{appendices}{
    \usepackage[titletoc,title]{appendix}
}{
}
\makeatother

\usepackage{cuted}

\newcommand{\sectionSI}[1]{%
  \refstepcounter{section}%
  \section*{SI Section \thesection\quad #1}%
  \addcontentsline{toc}{section}{SI #1}%
}

\usepackage{titletoc}
\newcommand{\myappendixtitle}{
    \twocolumn[
    \centering
    \bfseries\Large Supplementary Information: Guidelines for Augmentation Selection in\\ Contrastive Learning for Time Series Classification
    \bigskip
    ]
}

\usepackage{tablefootnote}

\usepackage[colorlinks=true, allcolors=blue]{hyperref}

\usepackage{bm}
\usepackage{cite}
\usepackage{amsmath,amssymb,amsfonts,mathrsfs}
\usepackage{algorithmic}
\usepackage{graphicx}
\usepackage{textcomp}
\usepackage{xcolor}
\def\BibTeX{{\rm B\kern-.05em{\sc i\kern-.025em b}\kern-.08em T\kern-.1667em\lower.7ex\hbox{E}\kern-.125emX}}

\usepackage{url}
\usepackage{booktabs} 
\usepackage{algorithmic}
\usepackage{comment}
\usepackage{colortbl}
\usepackage[ruled]{algorithm2e} 

\usepackage{multirow}
\usepackage{caption}
\usepackage{fancyhdr}

\usepackage{lscape}
\usepackage{pdflscape}
\usepackage[caption=false,font=normalsize,labelfont=sf,textfont=sf]{subfig}

\usepackage{pgfgantt}
\usepackage{graphicx}

\usepackage{makecell}

\newcolumntype{H}{>{\setbox0=\hbox\bgroup}c<{\egroup}@{}}

\begin{document}

\title{Guidelines for Augmentation Selection in Contrastive Learning for Time Series Classification}

\author{
Ziyu Liu$^1$
\and
Azadeh Alavi$^1$\and
Minyi Li \And
Xiang Zhang$^2$
\affiliations
$^1$RMIT, 
$^2$University of North Carolina, Charlotte\\
\emails
ziyu.liu2@student.rmit.edu.au,
\{azadeh.alavi, minyi.li2\}@rmit.edu.au, 
xiang.zhang@charlotte.edu
}

\author{
\IEEEauthorblockN{Ziyu Liu$^1$, 
Azadeh Alavi$^1$,
Minyi Li$^1$, Xiang Zhang$^2$}\\
\IEEEauthorblockA{
$^1$ RMIT, Australia, $^2$ UNC Charlotte, USA\\
Email: ziyu.liu2@student.rmit.edu.au\\
}
}

\maketitle

\begin{abstract}
Self-supervised contrastive learning has become a key technique in deep learning, particularly in time series analysis, due to its ability to learn meaningful representations without explicit supervision. 
Augmentation is a critical component in contrastive learning, where different augmentations can dramatically impact performance, sometimes influencing accuracy by over 30\%. However, the selection of augmentations is predominantly empirical which can be suboptimal, or grid searching that is time-consuming.
In this paper, we establish a principled framework for selecting augmentations based on dataset characteristics such as trend and seasonality. Specifically, we construct 12 synthetic datasets incorporating trend, seasonality, and integration weights. We then evaluate the effectiveness of 8 different augmentations across these synthetic datasets, thereby inducing generalizable associations between time series characteristics and augmentation efficiency.
Additionally, we evaluated the induced associations across 6 real-world datasets encompassing domains such as activity recognition, disease diagnosis, traffic monitoring, electricity usage, mechanical fault prognosis, and finance. These real-world datasets are diverse, covering a range from 1 to 12 channels, 2 to 10 classes, sequence lengths of 14 to 1280, and data frequencies from 250 Hz to daily intervals. 
The experimental results show that our proposed trend-seasonality-based augmentation recommendation algorithm can accurately identify the effective augmentations for a given time series dataset, achieving an average Recall@3 of 0.667, outperforming baselines. Our work provides guidance for studies employing contrastive learning in time series analysis, with wide-ranging applications.
\textit{All the code, datasets, and analysis results will be released at \url{https://github.com/DL4mHealth/TS-Contrastive-Augmentation-Recommendation} after acceptance.}

\end{abstract}

\begin{IEEEkeywords}
time series classification,
contrastive learning, augmentation,  signal decomposition, self-supervised learning
\end{IEEEkeywords}

\section{Introduction} 
\label{sec:introduction}


Self-supervised contrastive learning has emerged as a significant topic in deep learning, gaining attention across diverse fields, especially in time series analysis~\cite{zhang2024self,yang2022timeclr,wickstrom2022mixing,zhang2022self}. Contrastive learning algorithms focus on learning meaningful data representations by differentiating between similar (positive) and dissimilar (negative) data pairs, thus enabling feature extraction without explicit supervision~\cite{chen2020simple,verma2021towards}. This technique is especially relevant in time series analysis because time series data often contain complex temporal dependencies and patterns that are difficult to capture with traditional methods~\cite{eldele2021time,chen2020big}. Moreover, it's generally expensive to gain gold-standard labels for time series data, especially in healthcare, traffic, manufacturing, and environmental monitoring.

However, the success of the contrastive learning paradigm significantly depends on the quality and variety of augmentations applied to the data (Appendix Figure~\ref{fig:contrastive_framework})~\cite{liu2023self,chen2020simple}. Augmentation refers to the process of creating modified versions of the original data to generate positive pairs for training~\cite{demirel2024finding}. The goal is to produce different views of the same data that are still semantically meaningful, helping the model learn invariant features. These augmentations introduce necessary variations and perturbations to the data, pushing the model to learn invariant features crucial for subsequent tasks~\cite{poppelbaum2022contrastive}.

\begin{figure}
    \centering
    \includegraphics[width=\linewidth]{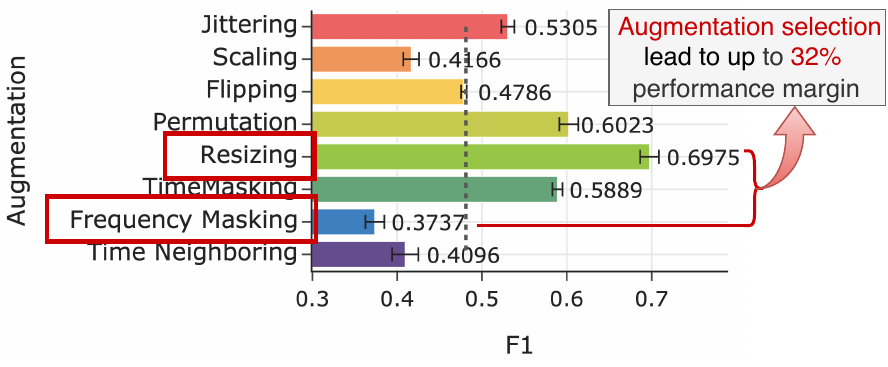}
    \vspace{-6mm}
    \caption{Augment selection can affect model performance dramatically (real-world fault detection dataset).
    }
    \label{fig:FD_bigmargin}
    \vspace{-6mm}
\end{figure}

Currently, the choice of augmentations largely follows an empirical approach, often based on prior research or initial experimental insights, which can be either time-consuming or suboptimal~\cite{zhang2022rethinking}. Furthermore, the \textit{performance of different augmentations} can vary dramatically across different datasets and tasks, sometimes leading to performance discrepancies as large as 32\% in accuracy due to the choice of augmentation alone (See \textit{Resizing} and \textit{Freq-Masking} in Fault Detection (FD) dataset, Figure~\ref{fig:FD_bigmargin}).
Moreover, the \textit{performance of the same augmentation} could vary significantly across datasets and tasks (Appendix Figure~\ref{fig:performance_real_bar}).

\begin{figure}
    \centering
    \includegraphics[width=0.86\linewidth]{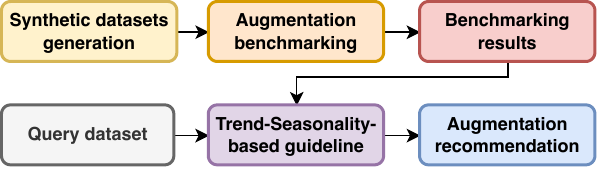}
    \caption{Study schematic}
    \label{fig:pip_overall}
    \vspace{-5mm}
\end{figure}

In this paper, we aim to provide principles for augmentation selection tailored to the specific characteristics of given time series datasets. We explore eight widely used augmentations in contrastive learning for time series classification. We methodically assess these augmentations across twelve synthetic datasets—designed to reflect the inherent trends and seasonality of time series data—and six real-world datasets (Section~\ref{sec:datasets}). Our objective is to establish a framework whereby the most effective augmentations are recommended based on the trend and seasonality of the dataset. 
We validate our recommendations through rigorous testing on six real-world datasets, ensuring that our findings are applicable universally across various time series data scenarios. We present the sketch of the working pipeline in Figure~\ref{fig:pip_overall}.

Through detailed empirical experiments and comparative analysis, this study endeavors to elucidate the impact of eight augmentations on the discriminative power and generalization ability of contrastive learning models tailored for time series classification. 
The insights derived from our research not only advance the current state-of-the-art in self-supervised learning but also provide practical guidance for applying contrastive learning techniques effectively in real-world settings.
This work marks a significant step toward a deeper understanding of the role augmentations play in contrastive learning, highlighting their critical importance in leveraging the untapped potential of time series data for diverse analytical tasks.

The contributions of this work are summarized below
\begin{itemize}
    \item We construct 12 synthetic time series datasets that cover linear and non-linear trends, trigonometric and wavelet-based seasonalities, and three types of weighted integration.

    \item We assess the effectiveness of 8 commonly used augmentations across all synthetic datasets, thereby elucidating the relationships between time series properties and the effectiveness of specific augmentations.
    
    \item We propose a trend-seasonality-based framework that precisely recommends the most suitable augmentations for a given time series dataset. Experimental results demonstrate that our recommendations significantly outperform those based on popularity and random selection.
\end{itemize}

\section{Related Work}
\label{sec:related_work}
 
\subsection{Contrastive learning in time series}
In time series analysis, where data often lacks explicit labels and exhibits complex temporal dependencies, contrastive learning offers a promising avenue for extracting informative features~\cite{zhang2024self}. By leveraging the sequential nature of time series data, self-supervised contrastive learning methods can capture long-range dependencies, temporal patterns, and underlying dynamics, facilitating various downstream tasks such as classification. Unlike supervised learning approaches that rely on labeled examples, contrastive learning aims to learn representations by maximizing the similarity between similar instances (positive pairs) while minimizing the similarity between dissimilar instances (negative pairs)~\cite{chen2020simple}. This process encourages the model to capture underlying structures and patterns present in the data without explicit supervision.

Augmentations, a significant component within the contrastive learning framework, are designed to introduce distortions and variations to the time series data, thereby enriching the learning process and promoting the discovery of invariant features~\cite{liu2023self}. 
The selection of the \textbf{augmentation} method is crucial for enhancing the quality and diversity of the learned representations, which largely affect the performance of contrastive learning algorithms~\cite{zhang2022rethinking}.

\subsection{Selection of augmentations in contrastive learning}

Differ from the fields of images and languages where intuitive guidelines can guide the creation of suitable augmented samples based on well-understood principles~\cite{miyai2023rethinking}, the process of manually selecting augmentations for time series data faces significant obstacles. This challenge arises from the complex and often imperceptible nature of the temporal structures in time series data, which makes it difficult to apply straightforward rules or human intuition effectively~\cite{luo2023time}.

The prevalent approach to selecting augmentations in time series analysis typically involves manual strategies, encompassing both direct and indirect empirical methods. Direct empirical selection entails conducting initial experiments to assess the effectiveness of widely used augmentations (Section~\ref{sub:augmentations}), a process that, despite its thoroughness, proves to be both time-consuming and resource-intensive. On the other hand, the indirect method relies on existing literature, adopting augmentations that are frequently utilized~\cite{ma2023survey}. However, the efficacy of such augmentations can vary significantly across different time series datasets, often rendering literature-based approaches less effective.



In this study, we benchmark the effectiveness of augmentations in time series datasets, offering guidance for applying augmentations to diverse datasets. A critical challenge we address is ensuring the applicability of our findings across a broad range of datasets. Our approach creates 12 synthetic datasets, leveraging signal decomposition to quantitatively evaluate the impact of eight augmentations. 
This process enables us to establish a bridge between \textit{the patterns of trend and seasonality within a dataset} and the \textit{effectiveness of specific augmentations}. We further validate our findings by applying these insights to six real-world datasets, thereby offering a more generalized and practical framework for augmentation selection in time series analysis.

\section{Contrastive Learning and Preliminaries}

\subsection{Self-Supervised contrastive learning framework}
Self-supervised contrastive learning learns effective representations by discriminating similarities and differences between samples. It aims to map similar samples closely while pushing dissimilar ones apart, based on augmented views~\cite{le2020contrastive}.
Contrastive learning is crucial for various reasons. First, contrastive learning eliminates the need to label data, relieving the burden of extensive annotation efforts, especially in some expertise-demand areas like healthcare~\cite{tang2020exploring, krishnan2022self}. Additionally, the approach of leveraging the intrinsic structure of data offers powerful and semantically meaningful embeddings that can remain stable and general across different tasks.  

The framework of self-supervised contrastive learning (Appendix Figure~\ref{fig:contrastive_framework}) typically includes the pre-training, fine-tuning, and testing stages. In the pre-training stage, a self-supervised encoder takes the positive pairs and negative pairs and learns to project the input samples into embeddings. Based on the learned embeddings, we calculate and minimize the contrastive loss, making positive pairs close to each other and negative pairs far away from each other in the embedding space. After pre-training, the well-trained encoder and a downstream classifier (depending on the fine-tuning mode) are used to make predictions for the input samples. In the fine-tuning stage, only a small amount of labeled samples are utilized to optimize the encoder and/or downstream classifier.
Typically, the training set in fine-tuning stage is a subset of the training set in per-taining stage. The ratio between these sets is called \textit{label ratio}, ranging from 0 to 1.
Lastly, the testing stage applies the fine-tuned encoder and classifier for new data.  


The principle behind effective contrastive learning is to \textit{uncover the shared information between similar instances} in the data. During pre-training, the unlabeled data are augmented, or in other words, transformed into variants of the original samples, which we call positive pairs~\cite{wang2024contrast}. Instead, negative pairs are samples from different classes or subjects, depending on the experimental design. 
The adopted encoder then projects the original sample and its variants (augmented identical samples) into feature representations. At this stage, the role of the contrastive loss function is to minimize the distance between positive pairs while maximizing the distance between negative pairs. Through training and optimization, the encoder learns to correctly distinguish between positive and negative pairs, which means that the augmented samples should have closer embedding with the original sample. 

Therefore, data augmentation methods (i.e., how to augment data) and the construction of contrastive pairs are the most important components of contrastive learning, as they will directly affect the performance of the encoder and limit the performance improvement.

\subsection{Augmentations in time series}
\label{sub:augmentations}
In self-supervised contrastive learning for time series, the augmentations can be seen as transformations that slightly alter the original sample to create a contrastive pair (along with the original sample) during the phase of pertaining. 

Suppose a time series sample $\bm{x} = \{x_1, x_2, \cdots, x_{N-1}, x_N \}$ contains $K$ timestamps while there is an observation at each timestamp. This work focuses on univariate time series (i.e., $x_N$ is a scalar), but our experimental design and conclusions can easily extend to multivariate time series (i.e., $x_N$ is a vector). 
Although there is a wide range of time series augmentations were proposed in the last three years since the emergence of self-supervised contrastive learning framework~\cite{chen2020simple}, some are borrowed bluntly from image processing (like rotation) or have limited usage (such as R-peak masking is limited to ECG signals, channel-wise neighboring only applies to multivariate time series). Therefore, in this work, we investigate 8 types of augmentations that are most commonly used in time series studies: jittering, scaling, flipping, permutation, resizing, time masking, and frequency masking.
For better intuition, we visualize the augmentations in Appendix Figure~\ref{fig:augmentations} (adopted from \cite{liu2023self}). The 8 augmentations we investigate in this work are elaborated in Appendix Section~\ref{SI:augmentations}.



\subsection{Signal decomposition}
Signal decomposition is a fundamental technique in time series analysis, where the goal is to dissect a time series into several interpretable components, typically including trend, seasonality, and residual component~\cite{theodosiou2011forecasting}. In math, a time series sample $\bm{x}$ can be decomposed into:
\begin{equation}
\label{eq:decompose}
    \bm{x}(t) = w_1 T(t) + w_2 S(t) + w_3 R(t)
\end{equation}
where $t \in \{1, 2, \cdots, K\}$ denotes the timestamp. The trend $T$, seasonality $S$, and residual component $R$ are functions of timestamp. The $w_1$, $w_2$, and $w_3$ are coefficients to adjust the scale of each component in the whole signal. 

The \textbf{trend} component reflects the underlying long-term progression of the dataset, showing how the data evolves over time, irrespective of cyclical or irregular patterns. 
\textbf{Seasonality} shows the pattern that repeats over a known, fixed period, such as daily, weekly, monthly, or quarterly seasonality. Identifying seasonality helps in understanding regular variations in the time series.
The \textbf{residual} component, also called noise in some studies, encompasses the random variation in the time series. These irregularities and fluctuations cannot be attributed to the trend or seasonality components.

The decomposition allows us to understand the underlying structure of the time series signal, facilitating better representation learning and downstream classification.

\subsection{Seasonal and Trend decomposition using Loess (STL)} 
STL is the most popular and effective method for decomposing a time series~\cite{cleveland1990stl}. It employs Loess, a local regression technique, to extract the trend and seasonality components, allowing STL to handle any type of seasonality pattern, not just fixed periodicity. STL's flexibility comes from its non-parametric nature, meaning it does not assume a specific statistical model for the time series data. This makes STL particularly useful for complex data with varying patterns of trend and seasonality.

In practice, STL decomposition iteratively fits loess smoother to the observed time series to separate the data into trend, seasonal, and residual components. This iterative approach ensures that the decomposition accurately reflects the underlying patterns of the time series, even in the presence of noise and outliers.


\begin{figure*}[ht]
    \centering
    \includegraphics[width=0.8\linewidth]{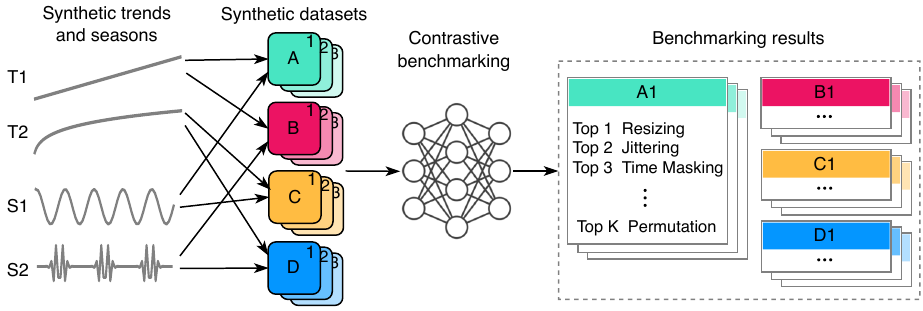}
    \caption{Generating synthetic datasets and benchmarking the augmentations. Based on two trends and two seasonalities, we have 4 dataset groups. In each dataset group, we generate 3 datasets based on the values of integration weights. Finally, we rank the augmentations for each dataset. }
    \label{fig:dataset_generation}
    \vspace{-3mm}
\end{figure*}

\section{Datasets}
\label{sec:datasets}

In this section, we introduce the process of building synthetic datasets to derive generalizable guidelines for augmentation selection. Inspired by STL decomposition, we construct 12 synthetic datasets by integrating trend, seasonal, and residual components. Additionally, we outline 6 real-world datasets to validate the guidelines learned from the synthetic datasets.

\subsection{Synthetic datasets generation}
\label{sec:synthetic_datasets}
In principle, an arbitrary time series signal can be decomposed into the sum of a trend, a seasonality, and a residual component, as shown in Eq.~\ref{eq:decompose}. The difference between the variety of time series datasets is the different functions of trend, seasonality, and residual parts, along with the weights when they integrate together. 

To increase the generalization of our results, therefore, we establish a series of synthetic datasets to cover as broad as time series patterns. To generate synthetic datasets, we need to consider 3 factors:
\begin{enumerate}
    \item How to select the functions of trend $T(t)$, seasonality  $S(t)$, and residual component $R(t)$?
    \item How to select the weights $w$ for each function when integrating them into the constructed time series sample?
    \item How to generate different classes in each synthetic dataset as we are working on classification tasks?
\end{enumerate}

Next, we answer the questions.

\textbf{Trends.} We investigate a linear trend $T_1(t)$ and a non-linear trend $T_2(t)$ as two representatives in time series data (Section~\ref{sec:discussion}):
\begin{equation}
    T_1(t) = \alpha t, \quad T_2(t) = t^{\alpha}
\end{equation}
where $\alpha$ is a coefficient to adjust the details of the trend. 

\textbf{Seasonalities.} Any time series signal can be regarded as the combination of a series of trigonometric functions, according to the Fourier series theorem~\cite{tolstov2012fourier}. This theorem states that any periodic signal can be represented as a sum of sines and cosines (trigonometric functions) with different frequencies, amplitudes, and phases. Thus, we employ a typical trigonometric function as $S_1(t)$:
\begin{equation}
\label{eq:S1}
    S_1(t) = \beta \textrm{sin}(\lambda t - \phi) + (1-\beta)\textrm{cos}(\lambda t - \phi)
\end{equation}
which is the weighted average of sine and cosine functions with phrase shift, where $\beta$, $\lambda$, and $\phi$ are parameters for amplitude, frequency, and phrase, respectively. 

Moreover, the Morlet Wavelet is a popular function for time-frequency analysis, particularly effective for analyzing non-stationary signals. We define $S_2(t)$ as:
\begin{equation}
    \label{eq:S2}
    S_2(t) =  \pi^{-\frac{1}{4}} e^{\beta tj}e^{- \frac{t^2}{2}}.
\end{equation}
This is a common form of simplified Morlet wavelet, which combines a complex exponential (sine wave) with a Gaussian window. In Eq~\ref{eq:S2}, $\beta = 2 \pi f_0$ where $f_0$ denotes the central frequency. Our implementation follows Scipy.

For the text below, we omit the $(t)$ in trend and seasonality functions for simplicity.

\textbf{Residuals components.}
To simplify matters, we use a standard Normal Gaussian distribution $\mathcal{N}(0, 1)$ as the residual component and keep it consistent across all synthetic datasets. 

\textbf{Dataset Generation.}
Considering 2 trends and 2 seasonalities, we generate 4 groups of datasets, naming Dataset groups A, B, C, and D, respectively.
For each dataset, we consider three situations: 
\begin{itemize}
    \item The trend is dominant ($w_1$ = 0.9, $w_2$ = 0.1).
    \item The trend and seasonality are even ($w_1$ = $w_2$ = 0.5).
    \item The seasonality is dominant ($w_1$ = 0.1, $w_2$ = 0.9).
\end{itemize}
For each synthetic dataset group, we build 3 datasets corresponding to the three situations above, denoted by a suffix `1', `2', or `3'.
Thus, we have 12 synthetic datasets in total, named A1, A2, ...., and D3. 
For example, dataset A1 is composed of linear trend and trigonometric seasonality, with $w_1 =0.9$.
We provide the workflow in Figure~\ref{fig:dataset_generation}.
See Table \ref{tab:synthetic_parameters} for more details. 

In this work, we focus on trend and seasonality, so keep \textbf{$w_3=0.3$} constant across all synthetic datasets.
We get 0.3 from preliminary studies: a larger residual component will overwhelm the synthetic dataset, making it too noisy; otherwise, a too small $w_3$ will make the datasets too simple, and all augmentations can achieve great performance, leading to less discriminative in benchmarking.

\paragraph{Generate classes in each dataset} Within each synthetic datasets, we use different coefficients in the trend and seasonality functions to simulate different classes, forming a 6-class classification task. In detail, we have two options for the $\alpha$ in trend $T$ (no matter $T_1$ or $T_2$): $\alpha = 0.2$ or $\alpha = -0.2$.  For Seasonality, the coefficient in seasonality has a larger effect on signals, for better generalization, we choose three options for the coefficient $\beta$ in $S1$:$\beta = 0.1$, $\beta = 0.5$, or $\beta = 0.9$. For $S2$, we use $\beta = 4$, $\beta = 5$, or $\beta = 6$.

By considering two options in $\alpha$ and three options in $\beta$, we have 6 classes in each of 12 datasets.

\paragraph{Generate samples in each class}
Within each class of each dataset, we need a large amount of samples to train deep learning models. In other words, we need intra-class sample variations. 
In principle, this variation can be synthesized by changing residual component, or changing the trends and seasonalities. While changing the residual component, the instinct properties of samples within a class are still obvious, making the classification performance very high, losing discriminative. Thus, we add perturbations to the trend and seasonality to formulate the sample variations.

In practice, we randomly select the alpha from two uniform distributions: $\alpha \in [0.1, 0.3]$, and $\alpha \in [-0.3, -0.1]$. The range $[0.1, 0.3]$ is calculated from the value 0.2 with a variation range 0.1, in other words, $\alpha = 0.2 \pm  0.1$ or $\alpha = -0.2 \pm  0.1$.
Simialrity, we randomly select beta from $\beta \in [0.05, 0.15]$, $\beta \in [0.45, 0.55]$, and $\beta \in [0.85, 0.95]$.
To have enough training power, we sampled 1000 samples for each class. 

In summary, we generated 12 synthetic datasets, each dataset contains 6 classes while each class has 1000 samples. The whole dataset cohort has 72,000 samples. We'll release the code for dataset synthetic, the users are free to increase the datasize, or change the setting to get their own datasets.

\begin{table*}[ht]
\centering
\caption{Synthetic datasets settings. 
We generate four dataset groups (\textit{A}, B, C, and D), and each group has three datasets (with suffixes of 1, 2, and 3, respectively). Each dataset has 6 classes. The $w_1$ is the weight of trend in the synthesized signal, correspondingly, the weight of seasonality (omitted here) is $w_2 = 1- w_1$.
}
\label{tab:synthetic_parameters}
\resizebox{\textwidth}{!}{
\begin{tabular}{@{}ccccccc|ccccccc@{}}
\toprule
\multirow{2}{*}{\textbf{Dataset}} & \multirow{2}{*}{\textbf{\begin{tabular}[c]{@{}c@{}}Decom-\\ position\\ weight \\ ($w_1$)\end{tabular}}} & \multicolumn{2}{c}{\textbf{Trend coefficient $\alpha$}} & \multicolumn{2}{c}{\textbf{Season coefficient $\beta$}} & \multirow{2}{*}{\textbf{Class}} & \multirow{2}{*}{\textbf{Dataset}} & \multirow{2}{*}{\textbf{\begin{tabular}[c]{@{}c@{}}Decom-\\ position\\ weight \\ ($w_1$)\end{tabular}}} & \multicolumn{2}{c}{\textbf{Trend coefficient $\alpha$}} & \multicolumn{2}{c}{\textbf{Season coefficient $\beta$}} & \multirow{2}{*}{\textbf{Class}} \\ \cmidrule(lr){3-6} \cmidrule(lr){10-13}

 &  & \multicolumn{1}{c|}{\makecell{$T_1 =$ \\ $\alpha t$}} & \multicolumn{1}{c|}{\makecell{$T_2 =$ \\ $t^{\alpha}$}} & \multicolumn{1}{c|}{\makecell{$S_1 =$ \\ $\beta\sin(\lambda t-\phi)+$ \\ $(1-\beta)\cos(\lambda t-\phi)+$ \\ $\beta\sin(\lambda t-\phi)$}} & \multicolumn{1}{c}{\makecell{$S_2 =$ \\ $\text{Morlet}(\beta,$ \\ $\phi(0,100))$}} & \multicolumn{1}{l|}{} & \multicolumn{1}{l}{} &  & \multicolumn{1}{c|}{\makecell{$T_1 =$ \\ $\alpha t$}} & \multicolumn{1}{c|}{\makecell{$T_2 =$ \\ $t^{\alpha}$}} & \multicolumn{1}{c|}{\makecell{$S_1 =$ \\ $\beta\sin(\lambda t-\phi)+$ \\ $(1-\beta)\cos(\lambda t-\phi)+$ \\ $\beta\sin(\lambda t-\phi)$}} & \multicolumn{1}{c}{\makecell{$S_2 =$ \\ $\text{Morlet}(\beta,$ \\ $\phi(0,100))$}} & \multicolumn{1}{l}{} \\ \midrule

\multirow{6}{*}{A1} & \multirow{6}{*}{0.1} & \multirow{3}{*}{0.1 $\sim$ 0.3} & \multicolumn{1}{c}{\multirow{3}{*}{- -}} & 0.05 $\sim$ 0.15 & - - & 0 & \multirow{6}{*}{C1} & \multirow{6}{*}{0.1} & \multicolumn{1}{c}{\multirow{3}{*}{- -}} & \multirow{3}{*}{0.1 $\sim$ 0.3} & - -  & 3.5 $\sim$ 4.5 & 0 \\
 &  &  & \multicolumn{1}{l}{} & 0.45 $\sim$ 0.55 & - - & 1 &  &  & \multicolumn{1}{l}{} &  & - -  & 4.5 $\sim$ 5.5 & 1 \\
 &  &  & \multicolumn{1}{l}{} & 0.85 $\sim$ 0.95 & - - & 2 &  &  & \multicolumn{1}{l}{} &  & - -  & 5.5 $\sim$ 6.5 & 2 \\
 &  & \multirow{3}{*}{-0.3 $\sim$ -0.1} & \multicolumn{1}{c}{\multirow{3}{*}{- -}} & 0.05 $\sim$ 0.15 & - - & 3 &  &  & \multicolumn{1}{c}{\multirow{3}{*}{- -}} & \multirow{3}{*}{-0.3 $\sim$ -0.1} & - -  & 3.5 $\sim$ 4.5 & 3 \\
 &  &  & \multicolumn{1}{l}{} & 0.45 $\sim$ 0.55 & - - & 4 &  &  & \multicolumn{1}{l}{} &  &  - - & 4.5 $\sim$ 5.5 & 4 \\
 &  &  & \multicolumn{1}{l}{} & 0.85 $\sim$ 0.95 & - - & 5 &  &  & \multicolumn{1}{l}{} &  & - -  & 5.5 $\sim$ 6.5 & 5 \\ \midrule
\multirow{6}{*}{A2} & \multirow{6}{*}{0.5} & \multirow{3}{*}{0.1 $\sim$ 0.3} & \multicolumn{1}{c}{\multirow{3}{*}{- -}} & 0.05 $\sim$ 0.15 & - - & 0 & \multirow{6}{*}{C2} & \multirow{6}{*}{0.5} & \multicolumn{1}{c}{\multirow{3}{*}{- -}} & \multirow{3}{*}{0.1 $\sim$ 0.3} & - -  & 3.5 $\sim$ 4.5 & 0 \\
 &  &  & \multicolumn{1}{l}{} & 0.45 $\sim$ 0.55 & - - & 1 &  &  & \multicolumn{1}{l}{} &  & - -  & 4.5 $\sim$ 5.5 & 1 \\
 &  &  & \multicolumn{1}{l}{} & 0.85 $\sim$ 0.95 & - - & 2 &  &  & \multicolumn{1}{l}{} &  &- -   & 5.5 $\sim$ 6.5 & 2 \\
 &  & \multirow{3}{*}{-0.3 $\sim$ -0.1} & \multicolumn{1}{c}{\multirow{3}{*}{- -}} & 0.05 $\sim$ 0.15 & - - & 3 &  &  & \multicolumn{1}{c}{\multirow{3}{*}{- -}} & \multirow{3}{*}{-0.3 $\sim$ -0.1} & - -  & 3.5 $\sim$ 4.5 & 3 \\
 &  &  & \multicolumn{1}{l}{} & 0.45 $\sim$ 0.55 & - - & 4 &  &  & \multicolumn{1}{l}{} &  &- -   & 4.5 $\sim$ 5.5 & 4 \\
 &  &  & \multicolumn{1}{l}{} & 0.85 $\sim$ 0.95 & - - & 5 &  &  & \multicolumn{1}{l}{} &  & - -  & 5.5 $\sim$ 6.5 & 5 \\ \midrule
\multirow{6}{*}{A3} & \multirow{6}{*}{0.9} & \multirow{3}{*}{0.1 $\sim$ 0.3} & \multicolumn{1}{c}{\multirow{3}{*}{- -}} & 0.05 $\sim$ 0.15 & - - & 0 & \multirow{6}{*}{C3} & \multirow{6}{*}{0.9} & \multicolumn{1}{c}{\multirow{3}{*}{- -}} & \multirow{3}{*}{0.1 $\sim$ 0.3} &  - - & 3.5 $\sim$ 4.5 & 0 \\
 &  &  & \multicolumn{1}{l}{} & 0.45 $\sim$ 0.55 & - - & 1 &  &  & \multicolumn{1}{l}{} &  & - -  & 4.5 $\sim$ 5.5 & 1 \\
 &  &  & \multicolumn{1}{l}{} & 0.85 $\sim$ 0.95 & - - & 2 &  &  & \multicolumn{1}{l}{} &  & - -  & 5.5 $\sim$ 6.5 & 2 \\
 &  & \multirow{3}{*}{-0.3 $\sim$ -0.1} & \multicolumn{1}{c}{\multirow{3}{*}{- -}} & 0.05 $\sim$ 0.15 & - - & 3 &  &  & \multicolumn{1}{c}{\multirow{3}{*}{- -}} & \multirow{3}{*}{-0.3 $\sim$ -0.1} & - -  & 3.5 $\sim$ 4.5 & 3 \\
 &  &  & \multicolumn{1}{l}{} & 0.45 $\sim$ 0.55 & - - & 4 &  &  & \multicolumn{1}{l}{} &  & - -  & 4.5 $\sim$ 5.5 & 4 \\
 &  &  & \multicolumn{1}{l}{} & 0.85 $\sim$ 0.95 & - - & 5 &  &  & \multicolumn{1}{l}{} &  & - -  & 5.5 $\sim$ 6.5 & 5 \\ \midrule
\multirow{6}{*}{B1} & \multirow{6}{*}{0.1} & \multicolumn{1}{c}{\multirow{3}{*}{- -}} & \multirow{3}{*}{0.1 $\sim$ 0.3} & 0.05 $\sim$ 0.15 & - - & 0 & \multirow{6}{*}{D1} & \multirow{6}{*}{0.1} & \multirow{3}{*}{0.1 $\sim$ 0.3} & \multicolumn{1}{c}{\multirow{3}{*}{- -}} &  - - & 3.5 $\sim$ 4.5 & 0 \\
 &  & \multicolumn{1}{l}{} &  & 0.45 $\sim$ 0.55 & - - & 1 &  &  &  & \multicolumn{1}{l}{} &  - - & 4.5 $\sim$ 5.5 & 1 \\
 &  & \multicolumn{1}{l}{} &  & 0.85 $\sim$ 0.95 & - - & 2 &  &  &  & \multicolumn{1}{l}{} &  - - & 5.5 $\sim$ 6.5 & 2 \\
 &  & \multicolumn{1}{c}{\multirow{3}{*}{- -}} & \multirow{3}{*}{-0.3 $\sim$ -0.1} & 0.05 $\sim$ 0.15 & - - & 3 &  &  & \multirow{3}{*}{-0.3 $\sim$ -0.1} & \multicolumn{1}{c}{\multirow{3}{*}{- -}} &  - - & 3.5 $\sim$ 4.5 & 3 \\
 &  & \multicolumn{1}{l}{} &  & 0.45 $\sim$ 0.55 & - - & 4 &  &  &  & \multicolumn{1}{l}{} & - -  & 4.5 $\sim$ 5.5 & 4 \\
 &  & \multicolumn{1}{l}{} &  & 0.85 $\sim$ 0.95 & - - & 5 &  &  &  & \multicolumn{1}{l}{} &  - - & 5.5 $\sim$ 6.5 & 5 \\ \midrule
\multirow{6}{*}{B2} & \multirow{6}{*}{0.5} & \multicolumn{1}{c}{\multirow{3}{*}{- -}} & \multirow{3}{*}{0.1 $\sim$ 0.3} & 0.05 $\sim$ 0.15 & - - & 0 & \multirow{6}{*}{D2} & \multirow{6}{*}{0.5} & \multirow{3}{*}{0.1 $\sim$ 0.3} & \multicolumn{1}{c}{\multirow{3}{*}{- -}} &  - - & 3.5 $\sim$ 4.5 & 0 \\
 &  & \multicolumn{1}{l}{} &  & 0.45 $\sim$ 0.55 & - - & 1 &  &  &  & \multicolumn{1}{l}{} &  - - & 4.5 $\sim$ 5.5 & 1 \\
 &  & \multicolumn{1}{l}{} &  & 0.85 $\sim$ 0.95 & - - & 2 &  &  &  & \multicolumn{1}{l}{} &  - - & 5.5 $\sim$ 6.5 & 2 \\
 &  & \multicolumn{1}{c}{\multirow{3}{*}{- -}} & \multirow{3}{*}{-0.3 $\sim$ -0.1} & 0.05 $\sim$ 0.15 & - - & 3 &  &  & \multirow{3}{*}{-0.3 $\sim$ -0.1} & \multicolumn{1}{c}{\multirow{3}{*}{- -}} &  - - & 3.5 $\sim$ 4.5 & 3 \\
 &  & \multicolumn{1}{l}{} &  & 0.45 $\sim$ 0.55 & - - & 4 &  &  &  & \multicolumn{1}{l}{} & - -  & 4.5 $\sim$ 5.5 & 4 \\
 &  & \multicolumn{1}{l}{} &  & 0.85 $\sim$ 0.95 & - - & 5 &  &  &  & \multicolumn{1}{l}{} & - -  & 5.5 $\sim$ 6.5 & 5 \\ \midrule
\multirow{6}{*}{B3} & \multirow{6}{*}{0.9} & \multicolumn{1}{c}{\multirow{3}{*}{- -}} & \multirow{3}{*}{0.1 $\sim$ 0.3} & 0.05 $\sim$ 0.15 & - - & 0 & \multirow{6}{*}{D3} & \multirow{6}{*}{0.9} & \multirow{3}{*}{0.1 $\sim$ 0.3} & \multicolumn{1}{c}{\multirow{3}{*}{- -}} &  - - & 3.5 $\sim$ 4.5 & 0 \\
 &  & \multicolumn{1}{l}{} &  & 0.45 $\sim$ 0.55 & - - & 1 &  &  &  & \multicolumn{1}{l}{} & - -  & 4.5 $\sim$ 5.5 & 1 \\
 &  & \multicolumn{1}{l}{} &  & 0.85 $\sim$ 0.95 & - - & 2 &  &  &  & \multicolumn{1}{l}{} &- -   & 5.5 $\sim$ 6.5 & 2 \\
 &  & \multicolumn{1}{c}{\multirow{3}{*}{- -}} & \multirow{3}{*}{-0.3 $\sim$ -0.1} & 0.05 $\sim$ 0.15 & - - & 3 &  &  & \multirow{3}{*}{-0.3 $\sim$ -0.1} & \multicolumn{1}{c}{\multirow{3}{*}{- -}} & - -  & 3.5 $\sim$ 4.5 & 3 \\
 &  & \multicolumn{1}{l}{} &  & 0.45 $\sim$ 0.55 & - - & 4 &  &  &  & \multicolumn{1}{l}{} & - -  & 4.5 $\sim$ 5.5 & 4 \\
 &  & \multicolumn{1}{l}{} &  & 0.85 $\sim$ 0.95 & - - & 5 &  &  &  & \multicolumn{1}{l}{} & - - & 5.5 $\sim$ 6.5 & 5 \\ \bottomrule
\end{tabular}
}
\end{table*}

\subsection{Real-world datasets}
Based on the results from 12 synthetic datasets, we can draw comprehensive conclusions about the effectiveness of various augmentations across a broad range of time series datasets. These datasets include those with linear or non-linear trends, those with trigonometric or Morlet seasonalities, and those that combine trends and seasonalities with varying weights.

To evaluate the correctness of the conclusions summarized on synthetic datasets, we evaluate the conclusions on 6 real-world datasets. The 6 datasets cover different classes (from 2 classes to 10 classes), from single channel to 12 channels, from short series (24) to long series (1280), and from diverse scenarios including human activity recognition, heart disease diagnosis, mechanical faulty detection, traffic pedestrian monitoring, electronic demand, and financial market prediction.
For datasets with unbalanced classes, we employ upsampling by randomly duplicating instances of the minority class until it matches the number of instances in the majority class, thereby balancing the sample amount in different classes in the training set.
We present the statistics of the dataset in Table~\ref{tab:real_world_datasets_statistics} and provide more details in Appendix Section~\ref{SI:realworld_datasets}.


\begin{table*}[ht]
\centering
\caption{Real-world datasets overview}
\label{tab:real_world_datasets_statistics}
\resizebox{\textwidth}{!}{%
\begin{tabular}{@{}ccccccccccc@{}}
\toprule
\multirow{2}{*}{\textbf{Dataset}} & \multirow{2}{*}{\textbf{\# class}} & \multirow{2}{*}{\textbf{\# channel}} & \multirow{2}{*}{\textbf{Length}} & \multicolumn{2}{c}{\textbf{Pre-train datasize}} & \multicolumn{3}{c}{\textbf{Fine-tune datasize}} & \multirow{2}{*}{\textbf{Frequency}} & \multirow{2}{*}{\textbf{Dataset description}} \\ \cmidrule(lr){5-9}
 &  &  &  & \textbf{Train} & \textbf{Val} & \textbf{Train} & \textbf{Val} & \textbf{Test} &  &  \\ \midrule
HAR & 6 & 3 & 206 & 5286 & 588 & 1762 & 1471 & 2947 & 50 Hz & Human activity recognition \\
PTB & 2 & 12 & 600 & 1440 & 160 & 480 & 200 & 200 & 250 Hz & Heart disease diagnosis (ECG) \\
FD & 3 & 1 & 1280 & 2008 & 224 & 669 & 2728 & 2728 & 16k Hz & \begin{tabular}[c]{@{}c@{}}Mechanical faulty Detection\end{tabular} \\
MP & 10 & 1 & 24 & 1024 & 114 & 341 & 2319 & 2319 & Per hour & \begin{tabular}[c]{@{}c@{}}Pedestrian volume \end{tabular} \\
ElecD & 7 & 1 & 96 & 8033 & 893 & 2677 & 7711 & 7711 & Per 2 mins & Electricity usage \\
SPX500 & 2 & 1 & 14 & 8973 & 998 & 2991 & 1246 & 1247 & Per day & Market indices \\ \bottomrule
\end{tabular}%
}
\vspace{-3mm}
\end{table*}

\section{Augmentation Benchmarking Methods}
\label{sec: methods}

\begin{figure*}[]
    \centering
    \includegraphics[width=0.72\linewidth]{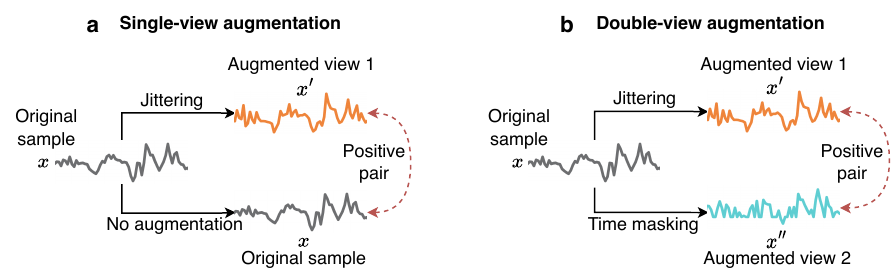}
    \caption{Example of single-view and double-view augmentation. 1) In single-view augmentation, we compare the augmented sample with the original sample; b) In double-view augmentation, we compare two augmented samples.}
    \label{fig:single_double}
\end{figure*}

In this section, we introduce how to compare the effectiveness of augmentations in contrastive learning frameworks. 
We start by identifying two concepts in augmentation: sing-view and double-view augmentations, as shown in Figure~\ref{fig:single_double}. 

\textit{Def 1.} \textbf{Single-view augmentation.} For an unlabeled time series sample $\bm{x}$, generated an augmented view $\bm{x}'$. Learns similar embeddings between \textit{the augmented view and the original view}: the positive pair is $(x, x')$.

\textit{Def 2.} \textbf{Double-view augmentation.} For an unlabeled time series sample $\bm{x}$, generated two augmented views $\bm{x}'$ and $\bm{x}''$. Learns similar embeddings between \textit{two augmented views}: the positive pair is $(x', x'')$. Note the two views could be generated with the same augmentation method (e.g., both using jittering but with different sampling/initialization).

\subsection{Single-view vs. single-view augmentation}
\label{sub:single_vs_single}

We first evaluate the effectiveness of single-view augmentations. For each of the 8 commonly-used time series augmentations introduced in Section~\ref{sub:augmentations}, we conduct 5 independent training (including pre-training and fine-tuning) in a single augmentation setting and report the average performance.
Note that in this work, we do not discuss combinational augmentations, such as jittering + flipping.
\subsection{Single-view vs. Double-view augmentation}

SimCLR and the follow-up studies use double-viewed augmentation in visual representation learning. However, in time series, single-view and double-view augmentations are both widely used and proven empirically helpful. 
Here we provide a comprehensive comparison between single-view and double-view augmentation.

In each dataset, we select the three most useful augmentations based on the results from single-view augmentation benchmarking (Section~\ref{sub:single_vs_single}), then conduct double-view augmentations. There are 6 situations in total (select 2 from 3 augmentations with replacement). Note, the top 3 augmentations and consequenced 6 situations are dataset-specific. 

\section{Trend-Seasonality-Based Recommendation}
\label{sec:trend_seasonality_based_recommendation}
The overarching aim of this work is to offer guidance on augmentation selection for any given time series dataset, which we refer to as the \textit{query dataset}. 
We introduce an innovative \textbf{trend-seasonality-based} method to recommend the most effective augmentations tailored to specific trend and seasonality patterns. Additionally, we propose a \textbf{popularity-based} recommendation as a proxy for augmentation selection from literature and include a \textbf{random recommendation} as a baseline for comparison.

\begin{figure*}
    \centering
    \includegraphics[width=0.8\linewidth]{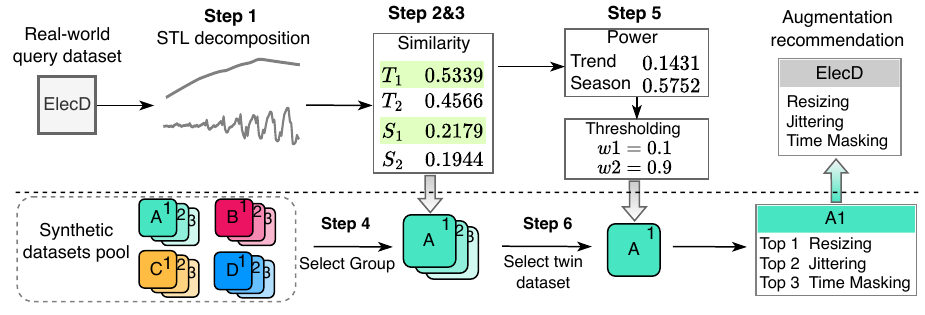}
    \caption{Workflow of the Trend-Seasonality-Based Recommendation System for Augmentations. Using the ElecD dataset as query dataset, for instance, we begin by decomposing it into trend \(T_q\) and seasonality \(S_q\) components using STL decomposition (Step 1). Next, we compare these components with synthetic trends and seasonalities (Steps 2-3). Based on the similarity in trends and seasonalities, we identify a closely related group of synthetic datasets, Group A (Step 4). We then calculate the power of \(T_q\) and \(S_q\) and set the values for weights \(w_1\) and \(w_2\) using thresholding (Step 5). This allows us to further refine Group A down to a specific dataset, A1 (Step 6). Knowing the most effective augmentations for A1 from our benchmarking results, we recommend these augmentations for the query dataset ElecD.}
    \label{fig:steps}
    \vspace{-3mm}
\end{figure*}

This trend-seasonality-based recommendation is the key novelty in this work. It personalizes recommendations to an arbitrary dataset based on the dataset properties. The key idea is to find a `synthetic twin dataset' for the query dataset, and then recommend the augmentations that are effective in the synthetic twin dataset.
The recommendation is conducted following the steps below, as shown in Figure~\ref{fig:steps}.

\noindent\textbf{Step 1}: Decompose the query dataset with STL, producing trend $T_q$ and $S_q$. The STL is sensitive to period size or the length of each seasonality. We address this issue in two manners: 1) if we can observe clearly seasonality (such as the heartbeat seasonality in ECG) or there exist meaningful seasonality (such as the weekly or bi-weekly cycle in finical time series), we will use the observed meaning seasonality to determine the period; 2) otherwise, we will try several periods separately based on the sampling frequency of the signal, then take the average of similarities.
For example, for a dataset with 256 HZ, we try the period window with 128, 64, 32, and 18, record the results, and take the average. 

\noindent\textbf{Step 2}: Calculate the similarity between $T_q$ and $T_1$ and $T_2$, respectively:

\begin{equation}
    \textrm{Sim}^T_{q,1} = d(T_q, T_1),
\end{equation}
\begin{equation}
    \textrm{Sim}^T_{q,2} = d(T_q, T_2)
\end{equation}
where $d$ is a similarity function (cosine similarity in this work). 

Likewise, we calculate the similarities in terms of seasonality, between $S_q$ and $S_1$ and $S_2$, respectively:
\begin{equation}
    \textrm{Sim}^S_{q,1} = d(S_q, S_1),
\end{equation}
\begin{equation}
    \textrm{Sim}^S_{q,2} = d(S_q, S_2)
\end{equation}

When calculating trend similarity, we constrain the $T_1$ and $T_2$ to have the same length as $T_q$. When calculating the seasonality similarity, we use a single period instead of the whole signal. For the period used in STL to decompose the query dataset, we adjust the length of $S_1$ and $S_2$ to align with one period of $S_q$. 

The calculated similarities for all real-world datasets are in Table~\ref{tab:similarity_power}.
Based on the similarity calculation, there would be three potential situations:
\begin{enumerate}
    \item If $T_q$ is closer to $T_1$, we will make recommendation considering $T_1$.
    \item If $T_q$ is closer to $T_2$, we will make recommendation considering $T_2$.
    \item If $T_q$ is in the middle between $T_1$ and $T_2$, we will not consider the influence of trend (i.e., neither $T_1$ nor $T_2$).
\end{enumerate}
The same rules apply to seasonality.  

\noindent\textbf{Step 3}: To determine the third case in Step 2, we further define the divergence score $\textrm{DS}_T$ that is used to describe the preference of $T_q$ on $T_1$ and $T_2$.
We calculate $\textrm{DS}_T$ through
\begin{equation}
    \textrm{DS}_T = 2 \frac{\textrm{Max}(\textrm{Sim}^T_{q,1}, \textrm{Sim}^T_{q,2}) - \textrm{Min}(\textrm{Sim}^T_{q,1}, \textrm{Sim}^T_{q,2})}{\textrm{Sim}^T_{q,1} + \textrm{Sim}^T_{q,2}}
\end{equation}

The numerator denotes the differences between $\textrm{Sim}^T_{q,1}$ and $\textrm{Sim}^T_{q,2}$. The denominator denotes the average between $\textrm{Sim}^T_{q,1}$ and $\textrm{Sim}^T_{q,2}$ (along with the coefficient of 2),  serving as a normalization scaler. By design, $\textrm{DS}_T \in [0, 2]$, the smaller $\textrm{DS}_T$ indicates the query trend is more neutral between $T_1$ and $T_2$. Based on empirical results, we set a threshold of $\textrm{DS}_T$ as 0.05: we don't consider the effect of the trend (only consider seasonality) if $\textrm{DS}_T < 0.05$, because in this case, the query trend is at the middle between Trend 1 and Trend 2. 

Similarly, we calculate divergence score $\textrm{DS}_S$ for seasonality and use the same threshold.

\noindent\textbf{Step 4}: Determine dataset group. Based on the calculation in Steps 2-3, we identify which trend and seasonality the query dataset is closer to. For example, if the query dataset is close to Trend 1 and Seasonality 2, the query dataset is close to the synthetic dataset group B. 

\noindent\textbf{Step 5}: Calculate the combination weights $w_1$ and $w_2$. After identifying the trend and seasonality the query dataset is close to, we calculate the power of trend and seasonality, respectively:

\begin{equation}
    P_T = \frac{1}{N} \sum_{i\leq N}{T_i^2}
\end{equation}
\begin{equation}
    P_S = \frac{1}{N} \sum_{i\leq N}{S_i^2}
\end{equation}
where $N$ denotes the length of the time series sample and $i$ denotes the timestamp. 

Dependent on the relative values of $P_T$ and $P_S$, we assign one of the three cases:
\begin{itemize}
    \item If $P_T/P_S \leq 5/9$, we set $w_1 = 0.1, w_2 = 0.9$;
    \item If $5/9 \leq P_T/P_S \geq 5$, we set $w_1 = 0.5, w_2 = 0.5$;
    \item If $P_T/P_S \geq 5$, we set $w_1 = 0.9, w_2 = 0.1$.
\end{itemize} 
The thresholds $5/9$ and $5$ are the middle points among $\frac{0.1}{0.9}$, $\frac{0.5}{0.5}$, and $\frac{0.9}{0.1}$. 
The calculated $P_T$ and $P_S$ for all real-world datasets are in Table~\ref{tab:similarity_power}. 

\noindent\textbf{Step 6}: Based on the weights $w_1$ and $w_2$, we further narrow down the synthetic twin dataset from dataset group to a specific synthetic dataset. For example, narrowing down from group B to dataset B2 if we got $w_1 = 0.5, w_2 = 0.5$.
The query dataset and the synthetic twin dataset are similar in terms of the trend, seasonality, and dominant component (taking $w_1$ and $w_2$ as proxy). Thus, we assume the effective augmentations in the synthetic twin dataset should also be effective in the query dataset. Therefore, the augmentation recommendation is simply recommending the top augmentations from the synthetic twin dataset (which we know from the benchmarking experiments).

To validate the effectiveness of our recommendation system, we compare it with two baselines: a random recommendation and a popularity-based recommendation. The latter is also developed by us and is calculated using benchmarking results from 12 synthetic datasets, it serves as a substitute for augmentation selection practices commonly found in the literature.

\subsection{Random Recommendation}

Random Recommendation refers to randomly selecting an augmentation from the augmentation pool without prior knowledge about the query dataset.

\subsection{Popularity-Based Recommendation} 
We first rank all the augmentations within each dataset, from Rank-1 to Rank-9 (there are 9 methods in total, including 8 augmentations plus a no-pretraining method). Then, we accumulate these rankings across all 12 synthetic datasets. As a result, augmentations with higher rankings are deemed more popular. Here, a higher ranking indicates a better performance or a smaller numerical rank (e.g., Top 1 is better than Top 2).
For any given query time series dataset, we recommend the top-K augmentations based on their popularity. 
Note that the recommended augmentations from popularity-based recommendations remain the same across any query dataset.

\begin{table}[t]
\caption{The trend and season similarity (Step 2) and power analysis (Step 5) for each real-world dataset. }
\label{tab:similarity_power}
\resizebox{\columnwidth}{!}{%
\begin{tabular}{c|ccc|ccc|cc}
\toprule
\multirow{2}{*}{\textbf{Dataset}} & \multicolumn{6}{c|}{\textbf{Average similarity of Trend and Seasonality}} & \multicolumn{2}{c}{\textbf{Average power}} \\ \cline{2-9} 
 & \textbf{T1} & \textbf{T2} & \textbf{$\textrm{DS}_T$} & \textbf{S1} & \textbf{S2} & \textbf{$\textrm{DS}_S$} & \textbf{Trend} & \textbf{Season} \\ \hline
HAR & 0.0511 & 0.0528 & 0.0329 & 0.263 & 0.0939 & 0.9479 & 0.8794 & 0.0933 \\
PTB & 0.1768 & 0.1585 & 0.1094 & 0.1814 & 0.3004 & 0.4938 & 0.2414 & 0.3990 \\
FD & 0.1031 & 0.0994 & 0.0365 & 0.0612 & 0.0575 & 0.0612 & 0.0177 & 0.5565 \\
MP & 0.2136 & 0.2037 & 0.0472 & 0.3615 & 0.3447 & 0.0476 & 0.7471 & 0.2466 \\
ElecD & 0.4271 & 0.3652 & 0.1562 & 0.2199 & 0.1814 & 0.1919 & 0.1431 & 0.5752 \\
SPX500 & 0.0273 & 0.023 & 0.1696 & 0.2832 & 0.2687 & 0.0526 & 0.9999 & 0.0003 \\  \bottomrule
\end{tabular}%
}
\vspace{-3mm}
\end{table}



\section{Results on Classification}
\label{sec:classification_results}
 
In this section, we first analyze the results from 12 synthetic datasets by dataset group (Section~\ref{sub:synthetic_results_1}), then report the results in terms of trend and seasonality (Section~\ref{sub:synthetic_results_2}), followed by the results from 6 real-world datasets (Section~\ref{sub:realworld_results}) for single-view versus single-view augmentation benchmarking. Additionally, we present a comparison of single-view and double-view augmentations.

We evaluate the performance of time series classification tasks with widely used metrics including accuracy, precision, recall, and F1 score. We mainly report F1 score in the text as it is a comprehensive metric balancing precision and recall. 
For a dataset, if two augmentations yield F1 scores that are closely comparable within a specified margin, we consider them to be equivalently effective. We define this margin as \textit{0.01 times the performance of no-pretraining}.

\begin{figure*}[t]
    \centering
    \includegraphics[width=0.95\textwidth]{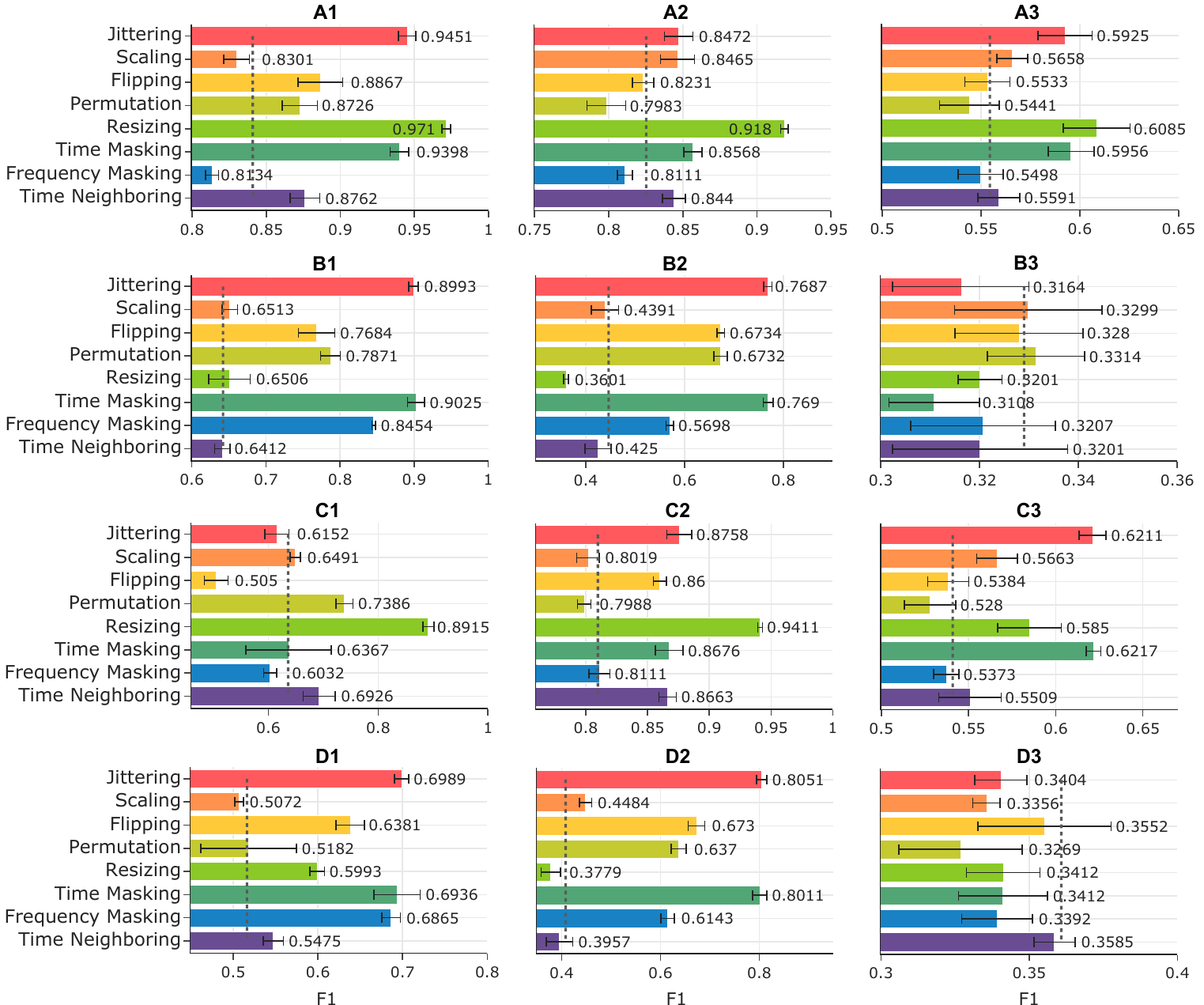}
    \caption{Results for single-view augmentation on synthetic datasets. The gray dashed line is the F1 score for no-pretraining. Note the x-axis scales vary across datasets.}
    \label{fig:performance_synthetic}
    \vspace{-3mm}
\end{figure*}

\subsection{Results of synthetic datasets by dataset groups}
\label{sub:synthetic_results_1}
We present the results for the 12 synthetic datasets by categorical groups represented by the prefixes A, B, C, and D. 
The datasets with the same prefix exhibit consistent trends and seasonality components, although the weights $w_1$ and $w_2$ differ. 
Due to space constraints in the main manuscript, we provide a detailed discussion exclusively only on the \textbf{top three} ranked augmentations. A comprehensive table containing results on all augmentations can be found in Table~\ref{tab:synthetic_single}.

\noindent\textbf{Dataset group A.}
\begin{itemize}
    \item We observed that 6, 6, and 5 augmentations respectively outperform the no-pretraining baseline on datasets A1, A2, and A3.

    \item Across all sub-datasets within Group A, \textbf{resizing} consistently emerges as the top-1 ranked augmentation, underscoring its effectiveness in synthetic datasets characterized by linear trends and trigonometric seasonalities. In addition to resizing, time masking and jittering also prove to be highly effective, securing the second and third ranks.
    \item The best F1 scores are 0.9715, 0.9184, and 0.6085, corresponding to increased margins of 13.05\%, 9.31\%, and 5.04\%, respectively. We observe a decreasing trend in the margin as the seasonality component becomes less dominant (with a lower weight of season), indicating that \textbf{the effectiveness of resizing is primarily driven by the compounded trigonometric seasonality.}
\end{itemize}

\noindent\textbf{Dataset group B.}
\begin{itemize}
    \item In datasets B1 and B2, 8 and 5 augmentations show superior performance compared to the no-pretraining baseline, respectively. However, in dataset B3, no augmentation over the baseline, only permutation and scaling show results close to (but lower than) the no-pretrain scenario.
    \item While Time masking and jittering share the same rank as the top augmentation for datasets B1 and B2, these augmentations did not demonstrate improvement over the no-pretraining baseline in dataset B3. This observation highlights that \textbf{time masking and jittering augmentations are particularly useful when the trigonometric seasonality predominates} in dataset group B.
    
    \item The best F1 scores are 0.9025, 0.769, and 0.3314, achieving margins of 26.04\%, 32.27\%, and 0.24\%, on datasets B1, B2, and B3, respectively. 
\end{itemize}

\noindent\textbf{Dataset group C.}
\begin{itemize}
    \item We observe 5, 6, and 5 augmentations outperform no-pretraining on datasets C1, C2, and C3, respectively.
    \item \textbf{Resizing} is the top augmentation for datasets C1 and C2, while for dataset C3 it occupies the third position. This indicates \textbf{the effectiveness of resizing weakens within group C as the weight of trigonometric seasonality decreases} and the weight of the non-linear trend increases. \textbf{This confirms the observation we get from Dataset group A.}

    \item For dataset C1, permutation and time-wise neighboring occupy the second and third positions, respectively. In dataset C2, jittering and time masking take the second and third ranks. However, for dataset C3, both time masking and jittering rank the top augmentation. 
    
    \item The highest F1 scores obtained in three datasets are 0.8915, 0.9411, and 0.6217, resulting in increases of 25.55\%, 13.13\%, and 8.08\% respectively compared to each no-pretraining baseline. \textbf{This observation is consistent with the tendency indicating that the effectiveness of resizing is associated with the weight of trigonometric seasonality.}
\end{itemize}

\noindent\textbf{Dataset group D.}
\begin{itemize}
    \item For datasets D1 and D2, 6 augmentations outperform the no-pretrain baseline, whereas the result of dataset D3 shows no augmentation achieving better performance. 
    
    \item Jittering and time masking emerged as the most effective augmentations, which rank first and second in dataset D1 and tied for first in dataset D2, but not in D3. This demonstrates that \textbf{jittering and time masking augmentations are only effective when the seasonality component plays a predominant or equal role} to the trend component. \textbf{We reach the same conclusion in Group B, even though it uses trigonometric seasonality, while here Morlet seasonality is employed.}
    In terms of the third-ranked augmentation, frequency masking is for dataset D1, while flipping is for dataset D2.
    
    \item The best F1 scores are 0.6989, 0.8051, and 0.3585 corresponding to increased margins of 18.27\%, 39.71\%, and -0.23\%, respectively.    
\end{itemize}

\begin{table*}[]
\caption{
Results on synthetic datasets for single-view augmentations comparison. 
If the performance of two augmentations is closer than a margin, which is adaptively defined as one percent of the no-pretraining F1 score, we consider them to perform equally, denoted by $\approx$.
The `Trig' is short the trigonometric functions.
}
\label{tab:synthetic_single}
\resizebox{\textwidth}{!}{%
\begin{tabular}{ccccclccc}
\toprule
\textbf{Data} & \textbf{Trend} & \textbf{Seasonality} & \textbf{\begin{tabular}[c]{@{}c@{}}Trend\\ $w_1$\end{tabular}} & \textbf{\begin{tabular}[c]{@{}c@{}}Seasonality\\ $w_2$\end{tabular}} & \multicolumn{1}{c}{\textbf{Single-view augmentation ranking}} & \textbf{\begin{tabular}[c]{@{}c@{}}Top-1 Aug\\ F1\end{tabular}} & \textbf{\begin{tabular}[c]{@{}c@{}}No-pretrain\\ F1\end{tabular}} & \textbf{\begin{tabular}[c]{@{}c@{}}Increased \\ Margin\end{tabular}} \\ \midrule
A1 & \multirow{3}{*}{Linear} & \multirow{3}{*}{Trig} & 0.1 & 0.9 & \begin{tabular}[c]{@{}l@{}}Resizing $$\textgreater$$ Jittering $\approx$ Time Masking $\textgreater$ Flipping \\ Time-neighboring $\approx$ Permutation $\textgreater$ No pretrain\end{tabular} & 0.9715 & 0.841 & 13.05\% \\ \cmidrule{6-6} 
A2 &  &  & 0.5 & 0.5 & \begin{tabular}[c]{@{}l@{}}Resizing $\textgreater$ Time Masking $\textgreater$ Jittering $\approx$  Scaling\\ $\approx$ Time-neighboring $\textgreater$ Flipping $\approx$ No pretrain\end{tabular} & 0.9184 & 0.8253 & 9.31\% \\ \cmidrule{6-6} 
A3 &  &  & 0.9 & 0.1 & \begin{tabular}[c]{@{}l@{}}Resizing $\textgreater$ Time Masking $\approx$ Jittering $\textgreater$ Scaling \\ $\textgreater$ Time-neighboring $\approx$ No pretrain\end{tabular} & 0.6085 & 0.5545 & 5.40\% \\ \midrule
B1 & \multirow{3}{*}{Linear} & \multirow{3}{*}{Morlet} & 0.1 & 0.9 & \begin{tabular}[c]{@{}l@{}}Time Masking $\approx$ Jittering $\textgreater$ Freq-Masking $\textgreater$\\ Permutation $\textgreater$ Flipping $\textgreater$ Scaling $\approx$ Resizing\\ $\textgreater$ Time-neighboring $\approx$ No pretrain\end{tabular} & 0.9025 & 0.6421 & 26.04\% \\ \cmidrule{6-6} 
B2 &  &  & 0.5 & 0.5 & \begin{tabular}[c]{@{}l@{}}Time Masking $\approx$ Jiitering $\textgreater$ Flipping $\approx$ Permutation \\ $\textgreater$ Freq-Masking $\textgreater$ No pretrain\end{tabular} & 0.769 & 0.4463 & 32.27\% \\ \cmidrule{6-6} 
B3 &  &  & 0.9 & 0.1 & Permutation $\approx$ No pretrain $\approx$ Scaling & 0.3314 & 0.329 & 0.24\% \\ \midrule
C1 & \multirow{3}{*}{\begin{tabular}[c]{@{}c@{}}Non-\\ linear\end{tabular}} & \multirow{3}{*}{Trig} & 0.1 & 0.9 & \begin{tabular}[c]{@{}l@{}}Resizing $\textgreater$ Permutation $\textgreater$ Time-neighboring $\textgreater$ \\ Scaling $\textgreater$ Time Masking $\approx$ No pretrain\end{tabular} & 0.8915 & 0.636 & 25.55\% \\ \cmidrule{6-6} 
C2 &  &  & 0.5 & 0.5 & \begin{tabular}[c]{@{}l@{}}Resizing $\textgreater$ Jittering $\textgreater$ Time masking $\approx$ Time-neighboring \\ $\approx$ Plipping $\textgreater$ Freq-Masking $\approx$ No pretrain\end{tabular} & 0.9411 & 0.8098 & 13.13\% \\ \cmidrule{6-6} 
C3 &  &  & 0.9 & 0.1 & \begin{tabular}[c]{@{}l@{}}Time masking $\approx$ Jittering $\textgreater$ Resizing $\textgreater$ Scaling $\textgreater$ \\ Time-neighboring $\textgreater$ No pretrain $\approx$ Flipping\\ $\approx$ Freq-Masking\end{tabular} & 0.6217 & 0.5409 & 8.08\% \\ \midrule
D1 & \multirow{3}{*}{\begin{tabular}[c]{@{}c@{}}Non-\\ linear\end{tabular}} & \multirow{3}{*}{Morlet} & 0.1 & 0.9 & \begin{tabular}[c]{@{}l@{}}Jittering $\textgreater$ Time Masking $\textgreater$ Freq-Masking $\textgreater$ Flipping $\textgreater$\\ Resizing $\textgreater$ Time-neighboring $\textgreater$ No pretrain $\approx$ Permutation\end{tabular} & 0.6989 & 0.5162 & 18.27\% \\ \cmidrule{6-6} 
D2 &  &  & 0.5 & 0.5 & \begin{tabular}[c]{@{}l@{}}Jittering $\approx$ Time Masking $\textgreater$ Flipping $\textgreater$ Permutation \\ $\textgreater$ Freq-Masking $\textgreater$ Scaling $\textgreater$ No pretrain\end{tabular} & 0.8051 & 0.408 & 39.71\% \\ \cmidrule{6-6} 
D3 &  &  & 0.9 & 0.1 & No pretrain $\approx$ Time-neighboring & 0.3585 & 0.3608 & -0.23\% \\ \bottomrule
\end{tabular}%
}
\end{table*}

\begin{table*}[]
\caption{Results on real-world datasets for single-view augmentation comparison. }
\label{tab:realworld_single}
\centering
\resizebox{\textwidth}{!}{%
\begin{tabular}{clccc}
\toprule
\textbf{Dataset} & \multicolumn{1}{c}{\textbf{Single-view augmentation ranking}} & \textbf{\begin{tabular}[c]{@{}c@{}}Top-1 Aug\\ F1\end{tabular}} & \textbf{\begin{tabular}[c]{@{}c@{}}No-pretrain\\ F1\end{tabular}} & \textbf{\begin{tabular}[c]{@{}c@{}}Increased \\ Margin\end{tabular}} \\ \midrule
HAR & \begin{tabular}[c]{@{}l@{}}Time Masking $\approx$ Jittering $\textgreater$ Freq-Masking $\textgreater$ Flipping  $\textgreater$ Scaling $\approx$ Time-neighboring $\textgreater$ No pretrain\end{tabular} & 0.8727 & 0.8391 & 3.36\% \\ \cmidrule{2-2}
PTB & \begin{tabular}[c]{@{}l@{}}Flipping $\approx$ Time Masking $\approx$ Time-neighboring $\approx$ Scaling $\approx$ Freq-Masking $\textgreater$ No pretrain\end{tabular} & 0.9707 & 0.9577 & 1.30\% \\ \cmidrule{2-2}
FD & \begin{tabular}[c]{@{}l@{}}Resizing $\textgreater$ Permutation $\textgreater$ Time Masking $\textgreater$ Jittering $\textgreater$ No pretrain $\approx$ Flipping\end{tabular} & 0.6975 & 0.4811 & 21.64\% \\ \cmidrule{2-2}
MP & \begin{tabular}[c]{@{}l@{}}Freq-Masking $\textgreater$ Scaling $\textgreater$ Flipping $\textgreater$ Time-neighboring $\textgreater$ Resizing $\textgreater$ Time Masking $\textgreater$ Jittering $\textgreater$ No pretrain\end{tabular} & 0.6008 & 0.4199 & 18.09\% \\ \cmidrule{2-2}
ElecD & \begin{tabular}[c]{@{}l@{}}Resizing $\textgreater$ Jittering $\approx$ Time-neighboring $\textgreater$ Time Masking $\approx$ Freq-Masking $\textgreater$ Scaling $\approx$ Flipping $\textgreater$ No pretrain\end{tabular} & 0.5535 & 0.5233 & 3.02\% \\ \cmidrule{2-2}
SPX500 & \begin{tabular}[c]{@{}l@{}}Flipping $\approx$ Resizing $\approx$ Time Masking $\approx$ Freq-Masking $\textgreater$ Jittering $\approx$ Scaling $\textgreater$\\ Time-neighboring $\textgreater$ Permutation $\textgreater$ No pretrain\end{tabular} & 0.5866 & 0.5712 & 1.54\% \\ \bottomrule
\end{tabular}%
}
\vspace{-3mm}
\end{table*}

\subsection{Results of synthetic datasets by trends and seasonalities}
\label{sub:synthetic_results_2}

In this part, we present the findings regarding the trends and seasonalities in the synthetic datasets and their relation to the top augmentations. We first report seasonality then trend.

\noindent\textbf{Seasonality component.}
Based on the experimental results from dataset groups with the same seasonality component, we can conclude that in these time series synthetic datasets, seasonalities hold greater importance compared to trends. In other words, \textbf{seasonality factors play a dominant role in determining the top augmentations and their performance.} More detailed observations are shown below:
\begin{itemize}
    \item Both dataset Groups A and C highlight \textbf{resizing} as the top augmentation, emphasizing the critical role of the compound trigonometric seasonality shared by these groups, despite differing in their trend components (linear versus non-linear).
    
    \item For Groups A and C, \textbf{increased margins consistently decrease as the influence of compound trigonometric seasonality diminishes} from A1 to A2 to A3, with similar observations noted for Group C. These findings correlate with the progressively reduced impact of the compound trigonometric seasonality.
    
    \item In Groups B and D, which utilize Morlet seasonality, the top three augmentations for sub-datasets with suffix 1 are consistently \textbf{time masking, jittering, and frequency masking}. 
    For sub-datasets with suffix 2, despite differences in their trend components, the leading augmentations remain \textbf{time masking, jittering, and flipping}.
    
    \item In Groups B and D, \textbf{the increased margins tend to increase as the influence of Morlet seasonality decreases}, indicating an association between seasonality type and augmentation effectiveness.
\end{itemize}





\noindent\textbf{Trend component.}
\begin{itemize}
    \item In dataset groups A and B with linear trends,\textbf{ we observe a decrease in both baseline and top augmentation performance as the weight of the trend component increases}. This suggests that the trend component may adversely affect the model's capability and the effectiveness of augmentations.

    \item In dataset groups C and D, which have non-linear trend components, the experimental results do not show consistent correlations. However, a pattern similar to that observed in the groups with linear trends emerges: when the trend component predominates in the time series samples ($w_1 = 0.9$), there is a significant deterioration in the performance of the no-pretraining baseline, with augmentations offering limited improvement.
\end{itemize}


\subsection{Results of real-world datasets}
\label{sub:realworld_results}

Next, we report the experimental results of single augmentation benchmarking on real-world datasets. The summarized detailed results can be found in Table~\ref{tab:realworld_single}.

\noindent\textbf{HAR.}
Time masking and jittering share the top augmentations, with frequency masking occupying the third-highest rank. The increased margin of top-1 augmentation compared to the no-pretraining baseline is 3.36\%.

\noindent\textbf{PTB.}
Five augmentations have similar performance, with the best augmentation achieving an increase of 1.30\% compared to the baseline performance. This dataset is not sensitive to augmentations.

\noindent\textbf{FD.}
Resizing, permutation, and time masking rank the top three augmentations, respectively. The resizing achieved an increased margin of 21.64\%, which is the highest margin across all real-world datasets, indicating this dataset is sensitive to augmentations. However, we didn't find reasonable causality for why FD is so sensitive to augmentations. 

\noindent\textbf{MP.}
Frequency masking, scaling, and flipping are the top three augmentations. The top augmentation improves the baseline by 18.09\%, making it the second-highest increased margin among the six real-world datasets. Moreover, it's worth mentioning that for the MP dataset, all eight benchmarked augmentations demonstrate improved performance, except permutation.

\noindent\textbf{ElecD.}
The top augmentation is resizing, providing a performance increase of 3.02\%, while time neighboring and jittering share the second rank. Although the increased margin may appear low and not noteworthy, similar to the situation with the MP dataset, for the ElecD dataset, all augmentations except permutation outperform the no-pretraining baseline.

\noindent\textbf{SPX500.}
Flipping, resizing, time masking, and frequency masking are all identified as the top augmentation (with similar performance) for this dataset, with a increased margin of 1.54\%. Furthermore, all eight augmentation methods achieve better performance compared to the no-pretraining baseline.

\begin{table}[t]
\centering
\caption{Results for single- vs. double-view augmentations for dataset group A. `N-p’ denotes no-pretraining baseline. The 'Single Best F1' shows the Top-1 Augmentation in the single-view setting, while the number in the parentheses indicates the increased margin achieved by the single-view augmentation.
}
\label{tab:single_double_A}
\resizebox{0.5\textwidth}{!}{
\begin{tabular}{c|c|c|cccc}
\toprule
Data & \begin{tabular}[c]{@{}c@{}}N-p\\ F1\end{tabular} & \begin{tabular}[c]{@{}c@{}}Single\\ Best F1\end{tabular} & \begin{tabular}[c]{@{}c@{}}Augmentation\\ 1\end{tabular} & \begin{tabular}[c]{@{}c@{}}Augmentation\\ 2\end{tabular} &\begin{tabular}[c]{@{}c@{}}Double\\ F1\end{tabular} & \begin{tabular}[c]{@{}c@{}}Increased\\ Margin\end{tabular} \\ \hline
\multirow{4}{*}{A1} & \multirow{4}{*}{0.841} & \multirow{4}{*}{\begin{tabular}[c]{@{}c@{}}0.9715\\ (13.05\%)\end{tabular}} & Resizing & Resizing & 0.918 & 7.70\% \\
 &  &  & Resizing & Jittering & 0.9638 & 12.28\% \\
 &  &  & Resizing & Time-masking & 0.9612 & 12.02\% \\
 &  &  & Jittering & Time-masking & 0.9343 & 9.33\% \\ \hline
\multirow{4}{*}{A2} & \multirow{4}{*}{0.8253} & \multirow{4}{*}{\begin{tabular}[c]{@{}c@{}}0.9184\\ (9.31\%)\end{tabular}} & Resizing & Resizing & 0.8704 & 4.51\% \\
 &  &  & Resizing & Time-masking & 0.8953 & 7.00\% \\
 &  &  & Resizing & Jittering & 0.8723 & 4.70\% \\
 &  &  & Time-masking & Jittering & 0.8356 & 1.03\% \\ \hline
\multirow{4}{*}{A3} & \multirow{4}{*}{0.5545} & \multirow{4}{*}{\begin{tabular}[c]{@{}c@{}}0.6085\\ (5.40\%)\end{tabular}} & Resizing & Resizing & 0.5462 & -0.83\% \\
 &  &  & Resizing & Time-Masking & 0.415 & -13.95\% \\
 &  &  & Resizing & Jittering & 0.4042 & -15.03\% \\
 &  &  & Time-Masking & Jittering & 0.5889 & 3.44\% \\ \bottomrule
\end{tabular}
}
\end{table}

\begin{table}[t]
\centering
\caption{Results for single- vs. double-view augmentations for dataset group B.}
\label{tab:single_double_B}
\resizebox{0.5\textwidth}{!}{
\begin{tabular}{c|c|c|cccc}
\toprule
Data & \begin{tabular}[c]{@{}c@{}}N-p\\ F1\end{tabular} & \begin{tabular}[c]{@{}c@{}}Single\\ Best F1\end{tabular} & \begin{tabular}[c]{@{}c@{}}Augmentation\\ 1\end{tabular} & \begin{tabular}[c]{@{}c@{}}Augmentation\\ 2\end{tabular} &\begin{tabular}[c]{@{}c@{}}Double\\ F1\end{tabular} & \begin{tabular}[c]{@{}c@{}}Increased\\ Margin\end{tabular} \\ \hline
\multirow{4}{*}{B1} & \multirow{4}{*}{0.6421} & \multirow{4}{*}{\begin{tabular}[c]{@{}c@{}}0.9025\\ (26.04\%)\end{tabular}} & Time-Masking & Time-Masking & 0.8937 & 25.16\% \\
 &  &  & Time-Masking & Jittering & 0.8858 & 24.37\% \\
 &  &  & Time-Masking & Freq-Masking & 0.8975 & 25.54\% \\
 &  &  & Jittering & Freq-Masking & 0.895 & 25.29\% \\ \hline
\multirow{4}{*}{B2} & \multirow{4}{*}{0.4463} & \multirow{4}{*}{\begin{tabular}[c]{@{}c@{}}0.769\\ (32.27\%)\end{tabular}} & Time-Masking & Time-Masking & 0.7008 & 25.45\% \\
 &  &  & Time-Masking & Jittering & 0.6717 & 22.54\% \\
 &  &  & Time-Masking & Flipping & 0.6221 & 17.58\% \\
 &  &  & Jittering & Flipping & 0.6342 & 18.79\% \\ \hline
B3 & 0.329 & \begin{tabular}[c]{@{}c@{}}0.3314\\ (0.24\%)\end{tabular} & Permutation & Permutation & 0.3391 & 1.01\% \\ \bottomrule
\end{tabular}
}
\vspace{-3mm}
\end{table}

\subsection{Results of single- vs. double-view augmentations}

\begin{figure}
    \centering
    \includegraphics[width=0.75\columnwidth]{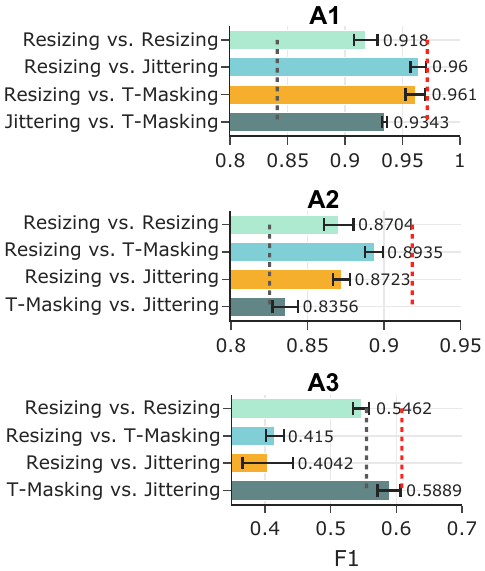}
    \caption{Results illustration for double-view augmentations of dataset group A. The grey and red dashed lines are F1 scores of no-pretraining and the best single-view, respectively. Full version of all dataset groups in Appendix Figure~\ref{fig:full_augset_bar}.}
    \label{fig:groupA_augset_bar}
    \vspace{-5mm}
\end{figure}

In this section, we compare the single-view augmentation method with double-view augmentations. Our goal is to assess whether a smaller distance within a positive pair (i.e., augmented view versus original sample) or a larger distance (i.e., augmented view versus another augmented view) is more effective for contrastive learning in time series classification.

\textbf{Experimental setting.} In practice, for each dataset, we first identify the three best augmentations, as shown in Figure~\ref{fig:performance_synthetic} and Table~\ref{tab:synthetic_single}. Let's call them Top-1, Top-2, and Top-3, respectively. 
For \textit{single-view augmentation}, the best performance is achieved by Top-1.
For \textit{double-view augmentation}, the results are symmetrical for the two views, leading to six possible combinations: (Top-1, Top-1), (Top-1, Top-2), (Top-1, Top-3), (Top-2, Top-2), (Top-2, Top-3), and (Top-3, Top-3). Given that Top-1 outperforms Top-2 and Top-3, we reasonably assume that the (Top-1, Top-1) combination is more effective than (Top-2, Top-2) and (Top-3, Top-3), allowing us to disregard the latter scenarios.
In summary, for double-view augmentations, we conduct four combinations: (Top-1, Top-1), (Top-1, Top-2), (Top-1, Top-3), and (Top-2, Top-3).

\textbf{Results.} 
We report the comparison results in Tables~\ref{tab:single_double_A}-\ref{tab:single_double_D}. We describe \textit{how to interpret the table} by taking the first row in Table~\ref{tab:single_double_A} for dataset A1 as an example: the no-pretraining (i.e., without any augmentation) F1 is 0.841; the best single-view augmentation boosts the F1 to 0.9715, claiming the increased margin of 13.05\%. We use the (Resizing, Resizing) combination for double-view augmentation (both $\bm{x}'$ and $\bm{x}''$ are generated by resizing but with different parameters), the double-view archives the F1 of 0.918, claiming 7.70\% margin over the no-pretraining baseline. So we know the single-view augmentation (0.9715) works better than the  (Resizing, Resizing) double-view augmentation (0.918).

In dataset group A, the results of all datasets show that \textbf{single-view augmentation can achieve a higher increased margin than double-view augmentation}. Both A1 and A2 show some improvement with double-view augmentations over no-pretraining baseline, but pairing the top one augmentation with itself does not consistently yield better performance compared to when it is paired with the second or third best augmentation method. 
For Dataset A3, only the (time masking, jittering) combination results in a slight improvement over the no-pretraining baseline, while other pairs present worse performance. This indicates that \textbf{double-view augmentations can sometimes lead to poorer results than using each augmentation individually}.


The results for Dataset Group B follow a similar pattern as observed in Groups A1 and A2 for B1 and B2. For Dataset B3, permutation is the only augmentation that outperforms the no-pretraining baseline, leading us to focus solely on the double-view of (permutation, permutation). This double-view augmentation obtains better performance (0.3391) than its single-view counterpart (0.3314). Although the performance of B3, at approximately 0.33, is not particularly high (likely due to the less discriminative nature of the linear trend across classes), it still surpasses that of random classification, which stands at 0.167 for a 6-class classification.

In Dataset C1, the double-view combination of (resizing, time-neighboring) performs equally to the single-view of resizing, with both achieving a margin of 25.55\%. For C2, single-view augmentation outperforms all double-view combinations. In the case of C3, the combination of (time-masking, time-masking) shows better performance than the top-ranked single-view. However, in all the datasets, the single-view and the best double-view augmentations perform similarly.

Dataset group D continues to follow the same pattern observed in dataset groups A and B, where single-view augmentation outperforms double-view augmentations.

In summary, in most cases, \textbf{single-view provides better performance}, while top-to-top dual-view enhancement fails to provide additional performance gains, not even surpassing single-view augmentation in most instances.

\begin{table}[t]
\centering
\caption{Results for single- vs. double-view augmentations for dataset group C.}
\label{tab:single_double_C}
\resizebox{0.5\textwidth}{!}{
\begin{tabular}{c|c|c|cccc}
\toprule
Data & \begin{tabular}[c]{@{}c@{}}N-p\\ F1\end{tabular} & \begin{tabular}[c]{@{}c@{}}Single\\ Best F1\end{tabular} & \begin{tabular}[c]{@{}c@{}}Augmentation\\ 1\end{tabular} & \begin{tabular}[c]{@{}c@{}}Augmentation\\ 2\end{tabular} &\begin{tabular}[c]{@{}c@{}}Double\\ F1\end{tabular} & \begin{tabular}[c]{@{}c@{}}Increased\\ Margin\end{tabular} \\ \hline
\multirow{4}{*}{C1} & \multirow{4}{*}{0.636} & \multirow{4}{*}{\begin{tabular}[c]{@{}c@{}}0.8915\\ (25.55\%)\end{tabular}} & Resizing & Resizing & 0.7973 & 16.13\% \\
 &  &  & Resizing & Permutation & 0.7952 & 15.92\% \\
 &  &  & Resizing & Time-Neighborning & 0.8915 & 25.55\% \\
 &  &  & Permutation & TIme-Neighborning & 0.7386 & 10.26\% \\ \hline
\multirow{4}{*}{C2} & \multirow{4}{*}{0.8098} & \multirow{4}{*}{\begin{tabular}[c]{@{}c@{}}0.9411\\ (13.13\%)\end{tabular}} & Resizing & Resizing & 0.8727 & 6.29\% \\
 &  &  & Resizing & Jittering & 0.9186 & 10.88\% \\
 &  &  & Resizing & Time-Masking & 0.9247 & 11.49\% \\
 &  &  & Jittering & Time-Masking & 0.8509 & 4.11\% \\ \hline
\multirow{4}{*}{C3} & \multirow{4}{*}{0.5409} & \multirow{4}{*}{\begin{tabular}[c]{@{}c@{}}0.6217\\ (8.08\%)\end{tabular}} & Time-Masking & Time-Masking & 0.6242 & 8.33\% \\
 &  &  & Time-Masking & Jittering & 0.6139 & 7.30\% \\
 &  &  & Time-Masking & Resizing & 0.6173 & 7.64\% \\
 &  &  & Jittering & Resizing & 0.6117 & 7.08\% \\ \bottomrule
\end{tabular}
}
\vspace{-3mm}
\end{table}

\begin{table}[t]
\centering
\caption{Results for single- vs. double-view augmentations for dataset group D.}
\label{tab:single_double_D}
\resizebox{0.5\textwidth}{!}{
\begin{tabular}{c|c|c|cccc}
\toprule
Data & \begin{tabular}[c]{@{}c@{}}N-p\\ F1\end{tabular} & \begin{tabular}[c]{@{}c@{}}Single\\ Best F1\end{tabular} & \begin{tabular}[c]{@{}c@{}}Augmentation\\ 1\end{tabular} & \begin{tabular}[c]{@{}c@{}}Augmentation\\ 2\end{tabular} &\begin{tabular}[c]{@{}c@{}}Double\\ F1\end{tabular} & \begin{tabular}[c]{@{}c@{}}Increased\\ Margin\end{tabular} \\ \hline
\multirow{4}{*}{D1} & \multirow{4}{*}{0.5162} & \multirow{4}{*}{\begin{tabular}[c]{@{}c@{}}0.6989\\ (18.27\%)\end{tabular}} & Jittering & Jittering & 0.5972 & 8.10\% \\
 &  &  & Jittering & Time-Masking & 0.6285 & 11.23\% \\
 &  &  & Jittering & Freq-Masking & 0.6143 & 9.81\% \\
 &  &  & Time-Masking & Freq-Masking & 0.6403 & 12.41\% \\ \hline
\multirow{4}{*}{D2} & \multirow{4}{*}{0.408} & \multirow{4}{*}{\begin{tabular}[c]{@{}c@{}}0.8051\\ (39.71\%)\end{tabular}} & Jittering & Jittering & 0.7529 & 34.49\% \\
 &  &  & Jittering & Time-Masking & 0.7496 & 34.16\% \\
 &  &  & Jittering & Flipping & 0.6536 & 24.56\% \\
 &  &  & Time-Masking & Flipping & 0.6533 & 24.53\% \\ \bottomrule
\end{tabular}
}
\vspace{-3mm}
\end{table}

\section{Results on augmentation recommendation}
\label{sec:recommendation_results}

\subsection{Evaluation metrics for Augmentation recommendation}
We evaluate the effectiveness of recommended augmentations by \textbf{Recall@K} which is borrowed from recommendation system studies~\cite{tang2021multisample}. 
Recall@K measures the proportion of relevant items found in the top-K recommendations provided by the model.
For example, Recall@3 in this work means that: how many among the recommended 3 best augmentations are truly in the 3 best augmentations for the query dataset.
\begin{equation}
\label{eq:recall@K}
    \textrm{Recall@K} = \frac{\mathcal{R}_K \bigcap \mathcal{T}_K}{K}
\end{equation}
where $\mathcal{R}_K$ denotes the set of recommended K best augmentations while $\mathcal{T}_K$ denotes the set of truly K best augmentations (Section~\ref{sub:realworld_results}).

In this study, we consider 8 augmentations, plus the no-pretraining baseline, there are 9 items for recommendation. In the main paper, we report Recall@1, Recall@2, and Recall@3, while presenting all the results (K=9) in the Appendix Table~\ref{tab:recall_at_k}.


\begin{table}[]
\centering
\caption{Evaluation of augmentation recommendation on real-world datasets (full version in Appendix Table~\ref{tab:recall_at_k}).}
\label{tab:recall_at_3}
\resizebox{0.5\textwidth}{!}{%
\begin{tabular}{cccccHccc}
\toprule
\multicolumn{2}{c}{\textbf{Recommendation method}} & \textbf{HAR} & \textbf{PTB} & \textbf{FD} & \textbf{MP*} & \textbf{ElecD} & \textbf{SPX500} & \textbf{Mean} \\ \midrule
\multirow{3}{*}{\textbf{Random}} & Recall@1 & 0.113 & 0.107 & 0.108 & 0.101 & 0.108 & 0.101 & 0.107 \\
 & Recall@2 & 0.235 & 0.217 & 0.222 & 0.216 & 0.218 & 0.209 & 0.22 \\
 & Recall@3 & 0.335 & 0.331 & 0.336 & 0.327 & 0.339 & 0.331 & 0.334 \\ \midrule
\multirow{3}{*}{\textbf{Popularity}} & Recall@1 & 1 & 0 & 0 & 0 & 0 & 0 & 0.2 \\
 & Recall@2 & 1 & 0 & 0 & 0 & 0.5 & 0 & 0.3 \\
 & Recall@3 & 0.667 & 0.333 & 0.667 & 0 & 0.667 & 0.667 & 0.6\\ \midrule
\multirow{3}{*}{\textbf{\begin{tabular}[c]{@{}c@{}}Trend-\\Seasonality\\ (Ours)\end{tabular}}}  & \cellcolor{lime!30} Recall@1 & \cellcolor{lime!30} 1 & \cellcolor{lime!30} 0 & \cellcolor{lime!30} 1 & \cellcolor{lime!30} 0 & \cellcolor{lime!30} 1 & \cellcolor{lime!30} 0 & \cellcolor{lime!30} \textbf{0.6} \\
 & \cellcolor{lime!30} Recall@2 & \cellcolor{lime!30} 1 & \cellcolor{lime!30} 0 & \cellcolor{lime!30} 1 & \cellcolor{lime!30} 0 & \cellcolor{lime!30} 1 & \cellcolor{lime!30} 0.5 & \cellcolor{lime!30} \textbf{0.7} \\
 & \cellcolor{lime!30} Recall@3 & \cellcolor{lime!30} 0.667 & \cellcolor{lime!30} 0.667 & \cellcolor{lime!30} 1 & \cellcolor{lime!30} 0 & \cellcolor{lime!30} 0.667 & \cellcolor{lime!30} 0.667 & \cellcolor{lime!30} \textbf{0.734} \\ \bottomrule
\end{tabular}%
}
\vspace{-3mm}
\end{table}

\subsection{Recommendation Results}
We compare three recommendation methods: our trend-seasonality-based, popularity-based, and random recommendations, as detailed in Section~\ref{sec:trend_seasonality_based_recommendation}. 
Each method generates a list, $\mathcal{R}_K$, of the \textit{recommended} top K augmentations.

On the six real-world datasets, we identify the \textbf{truly} top K augmentations, denoted as $\mathcal{T}_K$, based on the results in Section~\ref{sub:realworld_results}. We then calculate Recall@K using the formula in Eq~\ref{eq:recall@K}.
We report the results in Table~\ref{tab:recall_at_3}. We provide a concrete example below for better interpretation. Recall@3 = 0.667 = 2/3 means that: 2 out of 3 recommended augmentations fall within the true 3 best augmentations. 

Please note that while we use the popularity-based recommendation as a baseline, it is also one of our proposed methods because the popularity is calculated based on the experiment results we obtained from the synthetic datasets.


Before discussing the results in detail, we walk through the Trend-Season-based recommendation process again, using the real-world dataset \textbf{ElecD} as an example, to help readers better understand the method. 
We will clean and organize our implementation thoroughly and then release it to the public. \textbf{Readers can easily obtain the recommended augmentations for their datasets by calling our \textit{ts-arm} package function in Python (\textit{pip install ts-arm}).} The \textit{TS-ARM} denotes Time Series Augmentation Recommendation Method. 

The key idea is to determine which synthetic dataset ElecD is most similar to. To this end, we follow the six steps designed in Section~\ref{sec:trend_seasonality_based_recommendation}.

\noindent\textbf{Step 1:} Decompose each time series sample in dataset ElecD independently using STL. Then, we take the average of the decomposed trend and seasonality to get the overall trend $T_{ElecD}$ and seasonality $S_{ElecD}$.


\noindent\textbf{Step 2:} Calculate the similarity between $T_{ElecD}$ and $T_1$, $T_2$, as well as the similarity between $S_{ElecD}$ and $S_1$, $S_2$. The similarity values are presented in Table~\ref{tab:similarity_power}, which are 0.5339 and 0.4566 for $T_1$, $T_2$, along with 0.2179 and 0.1944 for $S_1$, $S_2$.

\noindent\textbf{Step 3:} Calculate the divergence score based on the similarity scores obtained in the previous step. We obtain a trend divergence of 0.1562 and a seasonality divergence of 0.1136, both exceeding the threshold of 0.05, which indicates that the similarities between the two trends/seasons are significant and cannot be ignored.

\noindent\textbf{Step 4:} Based on the calculations in Steps 2-3, $T_{ElecD}$ is similar to $T_1$ and $S_{ElecD}$ is similar to $S_1$. 
Therefore, we can identify that \textbf{Dataset group A} is the most similar synthetic dataset to dataset ElecD.

\noindent\textbf{Step 5:} Calculate the power of the trend and seasonality components of dataset ElecD, and we obtain $P_T^{ElecD}=0.1431$ and $P_S^{ElecD}=0.5752$.
Further, $P_T^{ElecD}/P_S^{ElecD} \leq 5/9$, we determine the decomposition weights of trend and season components to be $w_1 = 0.1$ and $w_2 = 0.9$.

\noindent\textbf{Step 6:} Based on the decomposition weights determined in Step 5, in dataset group A, we select \textbf{A1} as the synthetic twin dataset, and use augmentations from A1 (from Table~\ref{tab:synthetic_single}) to make recommendations for dataset ElecD. 

Next, we detail our recommendation process and the calculation of Recall@K as follows:
\begin{itemize}
    \item For users requiring only the best augmentation, we recommend the top augmentation from dataset A1, which is resizing. This matches the true top augmentation for dataset ElecD as shown in Table~\ref{tab:realworld_single}, resulting in a Recall@1 of 1.

    \item If the user needs two augmentations, we suggest the two best from A1: resizing and jittering. These correspond to the top two augmentations in ElecD, achieving a Recall@2 of 1.
    
    \item For recommendations involving three augmentations, we provide the three best from A1: resizing, jittering, and time masking. Two of these augmentations are among the top three for ElecD, leading to a Recall@3 of 0.667.
\end{itemize}


Next, let's extend to all real-world datasets. Among the six datasets, as shown in Table~\ref{tab:similarity_power}, HAR and FD have smaller trend divergence scores than the empirical threshold (0.05), so we only focus on seasonality. For MP, both the trend and seasonality divergence scores are smaller than the threshold, which means that MP is not similar to any of our trends or seasonalities: indicating the guidelines provided in this paper do not apply to MP. Thus, we calculate Recall@K for five datasets. 

As shown in Table~\ref{tab:recall_at_3}, our Trend-Season-based recommendation significantly outperforms both random and popularity-based methods. Although Popularity-based is obviously better than Random recommendation, \textbf{our method significantly surpasses the Popularity-based recommendation} by a great margin: 40\% absolute improvement in Recall@2 and 13.4\% in Recall@3.
In detail, when recommending the top augmentation, our method accurately matches the best augmentation in 3 out of 5 real-world datasets. In comparison, the popularity-based method achieves 1 match, and the random method has no matches.
Moreover, when recommending two augmentations, our method outperforms the other two baselines on 4 out of 5 datasets, with 3 achieving an exact match with the top two augmentations.
When recommending the three best augmentations, our trend-season-based method outperforms the random method, and is comparable with or exceeds the popularity-based recommendation across all datasets.

For completeness, we add the evaluation on MP in Appendix Table~\ref{tab:recall_at_k}. We observe that both popularity-based methods and our approach do not yield results in the top three recommendations, affirming the effectiveness of our divergence score. This points out a limitation: when the query dataset does not closely resemble any of the 12 synthetic datasets we used, the current guidelines become inapplicable. One promising future direction is to increase the coverage and generalizability. Further details are discussed in Section~\ref{sec:discussion}.

Overall, all experimental results robustly support the effectiveness of our Trend-Seasonality-based method in recommending suitable augmentations for a diverse range of time series datasets for contrastive learning.

\section{Discussion}
\label{sec:discussion}
In this work, we constructed 12 synthetic datasets based on signal decomposition rules, gained insights into the effectiveness of augmentations for time series analysis, and validated the conclusions on 6 diverse real-world datasets. We also introduced a trend-seasonality-based method to recommend the most effective augmentation for any given time series dataset. Next, we discuss the limitations of this study and outline several important directions for future research.

\textbf{More patterns of trends and seasonalities.}
In this study, we investigated two synthetic trends and two synthetic seasonalities, resulting in four combinations of trends and seasons.
However, in real-world scenarios, this may not suffice to cover the high diversity of time series data types. 
It's important to investigate trend and seasonality patterns to increase robustness in terms of data complexity and variability.
We propose three specific directions for improvement:
\begin{itemize}
    \item Investigating more complex trend and seasonality components to cover the diverse patterns in real-world datasets.
    \item Incorporating compound seasonality components by combining two or more functions of seasonalities (such as Morlet + Cosine), mirroring the complexity of seasonal patterns encountered in practical scenarios.
    \item Select highly representative real-world datasets as \textit{anchor datasets} for specific types of time series, akin to the concept of synthetic datasets used in this study. For instance, a large-scale high-quality ECG dataset could serve as the anchor dataset for other small-scale ECG data even though this dataset is monitoring a different disease with the anchor.
\end{itemize}

\textbf{More contrastive models.}
We focused exclusively on a standard contrastive framework derived from SimCLR, utilizing a Transformer as the backbone and implementing the Normalized Temperature-scaled Cross Entropy Loss (NT-Xent)~\cite{aagren2022nt}. 
It would be a valuable extension of this research to explore additional state-of-the-art contrastive frameworks, employing different backbones and loss functions.
%

\textbf{Alternative similarity metrics.}
In this study, we employed cosine similarity to calculate the similarity between real-world trend and seasonality to their synthetic counterparts. 
However, exploring alternative similarity measures such as Pearson similarity or self-learning approaches could potentially enhance the robustness of the results and provide additional insights.

\textbf{Divergence score thresholding.} In this work, we empirically select the threshold of divergence score (Step 3). This is mainly limited due to testing on only six real-world datasets, making it challenging to draw definitive conclusions about thresholding. As we expand our analysis to include more trends and seasonal variations, it will be essential to evaluate our methods across a broader and more diverse range of real-world datasets. This approach will enable us to establish more robust thresholds for the divergence score. Additionally, we anticipate developing a more effective method for calculating the divergence score in the future.

\textbf{More results analysis.} 
Our experiments on 8 single-view augmentations, 4 double-view augmentations across 12 synthetic datasets and 6 real-world datasets, along with 3 different recommendation methods, yielded a substantial volume of results. 
In Sections~\ref{sec:classification_results}-\ref{sec:recommendation_results}, we tried to present as much information as possible in our analysis of the results. Nevertheless, there remain aspects that we cannot fully explain or from which we cannot draw consistent conclusions.

For example, we notice that FD and MP, among all datasets, are most sensitive to augmentations, where augmentation outperforms no-pretraining by 21\% and 18\% respectively. However, we cannot find any common properties between them, in terms of channel numbers, class numbers, sample length, sampling rate (Table~\ref{tab:real_world_datasets_statistics}), decomposed trend or seasonalities (Table~\ref{tab:similarity_power}). Thus, we have to leave this as an opening issue for future work. 

\section{Conclusion} 
\label{sec:conclusion}

Through comprehensive analysis of 12 originally established synthetic datasets and the evaluation of 8 common augmentations, we have identified critical associations that guide the selection process. Our trend-seasonality-based recommendation system precisely tailors augmentation suggestions to specific dataset characteristics, consistently outperforming popularity-based and random recommendations. These findings not only enhance our understanding of contrastive learning dynamics but also pave the way for more effective implementations in practical settings across various industries.

\section*{Acknowledgement}
Z. L. is supported by the Australian Government Research Training Program (RTP) Scholarship at RMIT, Australia.
This work is partially supported by the National Science Foundation under Grant No. 2245894. Any opinions, findings, conclusions or recommendations expressed in this material are those of the authors and do not necessarily reflect the views of the funders.

\bibliographystyle{IEEEtran}
\bibliography{benchmarking_ts} 


\clearpage
\myappendixtitle

\newcounter{sectionSI}

\setcounter{figure}{0}
\setcounter{table}{0}

\sectionSI{Details of Augmentations in Time Series}
\label{SI:augmentations}

\begin{itemize}
    \item \textbf{Jittering}, the process of integrating random noise into the input sample, stands out as a widely favored augmentation technique due to its simplicity and efficacy~\cite{sarkar2021detection}. This method enhances a time series dataset by creating a modified version of the original sample, denoted as {$\bm{x}'$}, through the addition of random noise to {$\bm{x}'$}. The distribution of noise (like Gaussian, Poisson, or Exponential) is chosen based on the dataset's specific attributes and the inherent nature of the noise. Among these, Gaussian noise is the most frequently employed option, recognized for its versatility and effectiveness in a broad range of applications.

    \item \textbf{Scaling} refers to adjusting the amplitude of the original sample, effectively modifying its range~\cite{han2021semi}. For instance, if a sample's amplitude ranges from $[-1,1]$, applying a scaling factor of $1.5$ alters the range to $[-1.5, 1.5]$. It's important to note that the scaling factor can vary at different time steps within the same sample and across various samples, enhancing the augmented dataset's diversity and resilience to diverse amplitude variations.
    
    \item \textbf{Flipping} in time series analysis involves inverting the sequence's time steps, essentially reversing the data's chronological order~\cite{sarkar2021detection}. Mathematically, for a series $\bm{x} = \{x_1, x_2, \cdots, x_{N-1}, x_N \}$, the process results in a flipped series $\bm{x}' = \{x_N, x_{N-1}, \cdots, x_2, x_1 \}$, reversing the elements' order in the time series.
    
    \item \textbf{Permutation} of a time series incorporates two main steps: 1) segmentation, which divides the series into multiple subsequences; and 2) permuting, which entails randomly rearranging the subsequences~\cite{jiang2021self}. Each subsequence represents a continuous portion of the original series. This technique is particularly useful when the precise order of data points is not crucial, but maintaining the overall data distribution is essential.

    \item \textbf{Resizing} contains two actions: 1) slicing (a.k.a. cropping) and 2) resizing. In the slicing process, a segment of the original time series is randomly selected to serve as the augmented sample, effectively shortening the series by removing a set number of steps from the end~\cite{chen2021clecg}. For example, from an original series $\bm{x} = \{x_1, x_2, \cdots, x_{N-1}, x_N \}$, a cropped version might look like $\bm{x}' = \{x_1, x_2, \cdots, x_{K-m} \}$, where $m$ represents the count of time steps excluded. To compensate for the reduction in series length post-slicing, resizing is employed to adjust the cropped sample back to the original length. This is achieved through techniques that do not alter the series' amplitude but change its length, such as compression and stretching~\cite{mehari2022self}. Compression might involve downsampling the series to half its original length by selecting every other data point, resulting in $\bm{x}' = {x_1, x_3, x_5, \cdots, x_{N-2}, x_N }$. Conversely, stretching entails lengthening the series through interpolation, filling in gaps with averages of adjacent data points to preserve the overall trend without introducing significant amplitude changes.

    \item \textbf{Time masking} drops out certain observations within the time series to augment the data. Various approaches to masking include covering a consecutive segment of the series (subsequence masking) or hiding individual points at random (random masking). The concealed values can either be replaced with zero (zero-masking) or substituted with a different value (rescale-masking), making it a widely adopted augmentation strategy~\cite{han2021semi}.
    
    \item \textbf{Frequency masking} follows a similar principle to time masking, but operates in the frequency domain rather than the time domain. For a given time series sample, we first convert it into a frequency spectrum using techniques like the Fast Fourier Transform (FFT), then perturb well-designed frequency components. Employing zero-masking and subsequence masking within this domain effectively acts as a filter, whether it be low-pass, band-pass, or high-pass~\cite{zhang2022self}.
    
    \item \textbf{Time-wise neighboring.} considers two temporally close samples as a positive pair, under the assumption that their temporal features remain relatively stable. 
    For a long time series $\widehat{\bm{x}} = \{x_1, x_2, \cdots, x_{2N-1}, x_{2N} \}$, segmenting this into two non-overlapping samples each of length $K$ yields $\bm{x} = \{x_1, x_2, \cdots, x_{N-1}, x_N \}$ and $\bm{x}' = \{x_{N+1}, x_{N+2}, \cdots, x_{2N-1}, x_{2N} \}$ as a positive pair. Conversely, a negative pair would include $\bm{x}$ and another sample not temporally adjacent to $\bm{x}$~\cite{tonekaboni2021unsupervised}.
\end{itemize}

\sectionSI{Details of real-world datasets}
\label{SI:realworld_datasets}


\textbf{HAR.}
Human Activity Recognition (HAR) dataset~\cite{misc_human_activity_recognition_using_smartphones_240} contains recordings of 30 health subjects performing daily tasks: waking, walking upstairs, walking downstairs, sitting, standing, laying. 
The 3-axial linear acceleration
during the tasks are recorded by a smartphone at 50Hz.  
Following the setting in~\cite{zhang2022self}, we divided the dataset into three parts: the pre-training set (58\%, 5,881 samples), the validation set (14\%, 1,471 samples), and the test set (28\%, 2,947 samples).
During the pre-training stage, to verify the model's convergence, we used 90\% of the pre-training set (5,286 samples) for training and the remaining 10\% (588 samples) for validation.
In the fine-tuning stage, we used a subset of the pre-training set, 30\% in this study, as the training set. Note, the labels of training set in fine-tuning stage are required.

\textbf{PTB.}
Physikalisch-Technische Bundesanstalt (PTB) dataset measures 12-lead ECG signals~\cite{bousseljot1995nutzung, goldberger2000physiobank}. To test the ability of contrastive learning in small-scale datasets, we take a subset of PTB and make it a binary task (Myocardial infarction vs. Healthy). We take 28 subjects to keep the dataset balanced. The selected subject ID and constructed subset will be publicly released in our GitHub repository. 

\textbf{FD.}
The Faulty Detection (FD) dataset~\cite{lessmeier2016condition} comprises condition monitoring data of bearing damage in electromechanical drive systems.
The signals are labeled by 3 classes according to bearing damage: undamaged, inner damaged, and outer damaged. We adopt the preprocessed data from~\cite{zhang2022self} to downsample the window length of 5, 120 to 1,280 for higher computational efficiency.

\textbf{MP.}
The Melbourne Pedestrian (MP) dataset is sourced from the UCR Time Series Classification Archive~\cite{UCRArchive2018}. 
This dataset measures pedestrian volume at 10 locations, categorized into 10 classes. Each trial consists of 24 values, each measuring the hourly volume and spanning a full 24-hour day.
We use the original dataset's training and test sets as provided. We treat the validation set as equivalent to the test set.

\textbf{ElecD.}
The ElectricDevices (ElecD) dataset is also obtained from the UCR Time Series Classification Archive~\cite{UCRArchive2018}.
It measures household electricity usage. The data are sampled from 251 households at two-minute intervals for a month. The dataset consists of 7 classes, each corresponding to 7 levels of consumers' pro-environmental behavior. 
The original dataset provides only the training and test sets, which we use directly. In this paper, we also use the test set as validation set.

\textbf{SPX500.}
The S\&P 500 (SPX500) dataset comprises the market index of 500 leading companies~\cite{spx500yahoo}. We utilize data from 1989-12-29 to 2024-03-01 and take two weeks of data as features to predict the daily return for the following day. Daily returns are calculated by the price difference between the closing and opening prices, and encoded into binary classes (1 denotes profit; 0 denotes loss). We split the training, validation, and test set in an 8:1:1 ratio. 

\begin{table}[ht]
\centering
\caption{Synthetic datasets overview}
\label{tab:my-table}
\resizebox{\columnwidth}{!}{%
\begin{tabular}{@{}ccc@{}}
\toprule
Dataset prefix & Trend component & Seasonality component \\ \midrule
A & Linear &  Trigonometric function \\
B & Linear & Morlet \\
C & Nonlinear &  Trigonometric function \\
D & Nonlinear & Morlet \\ \bottomrule
\end{tabular}%
}
\end{table}

\begin{figure*}[]
    \centering
    \includegraphics[width=0.65\textwidth]{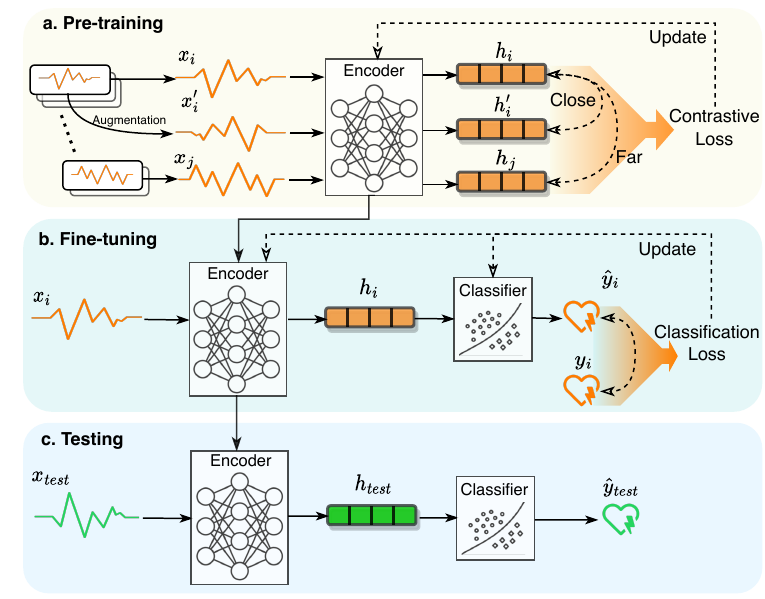}
    \caption{Framework for contrastive learning from~\cite{liu2023self}. Pipeline of self-supervised contrastive learning which is composed of three stages. \textbf{a.} The pre-training receives \textit{unlabelled} time series sample $\bm{x}_i$ as anchor sample, the augmented sample $\bm{x}'_i$ as positive sample while a different sample $\bm{x}_j$ as negative sample. The $\bm{h}_i$, $\bm{h}'_i$ and $\bm{h}_j$ denotes learned embedding of the original sample $\bm{x}_i$, positive pair $\bm{x}'_i$, and negative pair $\bm{x}_j$, respectively. 
A contrastive loss is calculated based on the distance among embeddings of samples, which is used to update the encoder through backpropagation. \textbf{b.} The well-trained encoder will be inherited by the fine-tuning stage which receives a \textit{labeled} sample and makes a prediction through a downstream classifier. A standard supervised loss function (e.g., cross-entropy) will be used to update the encoder and/or classifier. \textbf{c.} The testing stage makes prediction based on the learned embedding $\bm{h}_{test}$ of an unseen test sample $\bm{x}_{\textrm{test}}$. 
    }
    \label{fig:contrastive_framework}
\end{figure*}

\begin{figure*}
    \centering
    \includegraphics[width=\linewidth]{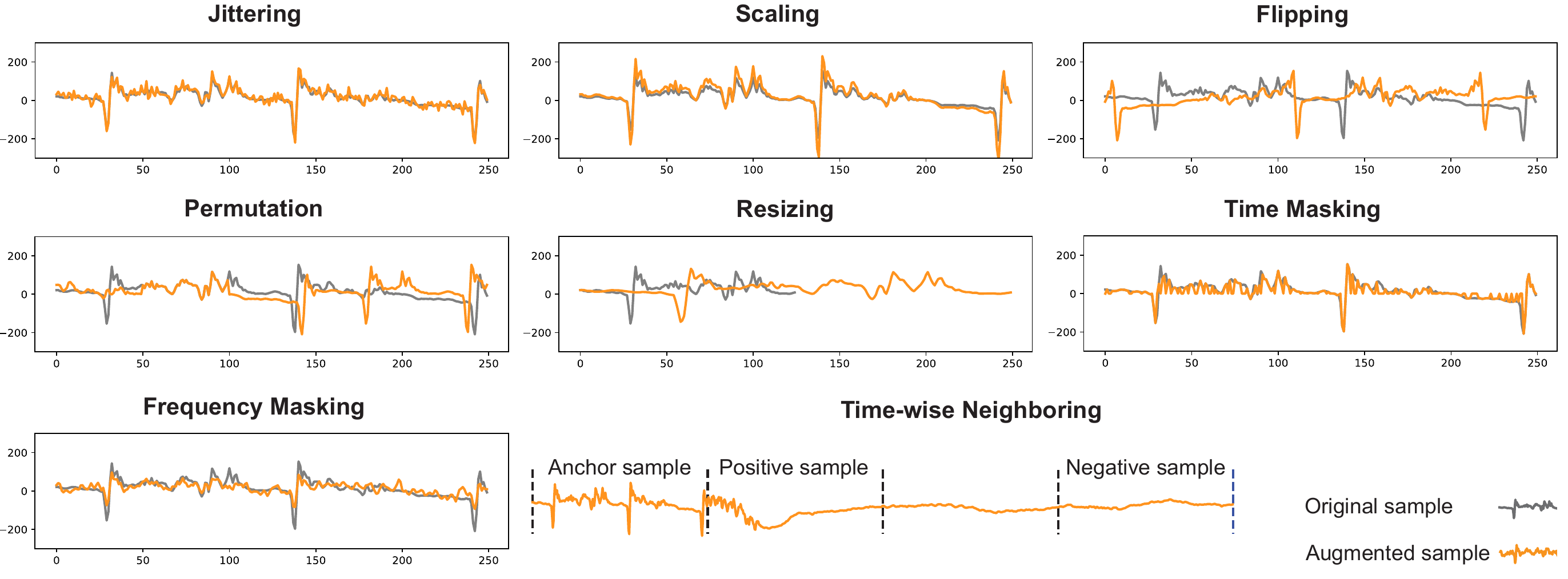}
    \caption{Examples of augmentation methods used in benchmarking.}
    \label{fig:augmentations}
\end{figure*}

\begin{figure*}[]
    \centering
    \includegraphics[width=0.9\linewidth]{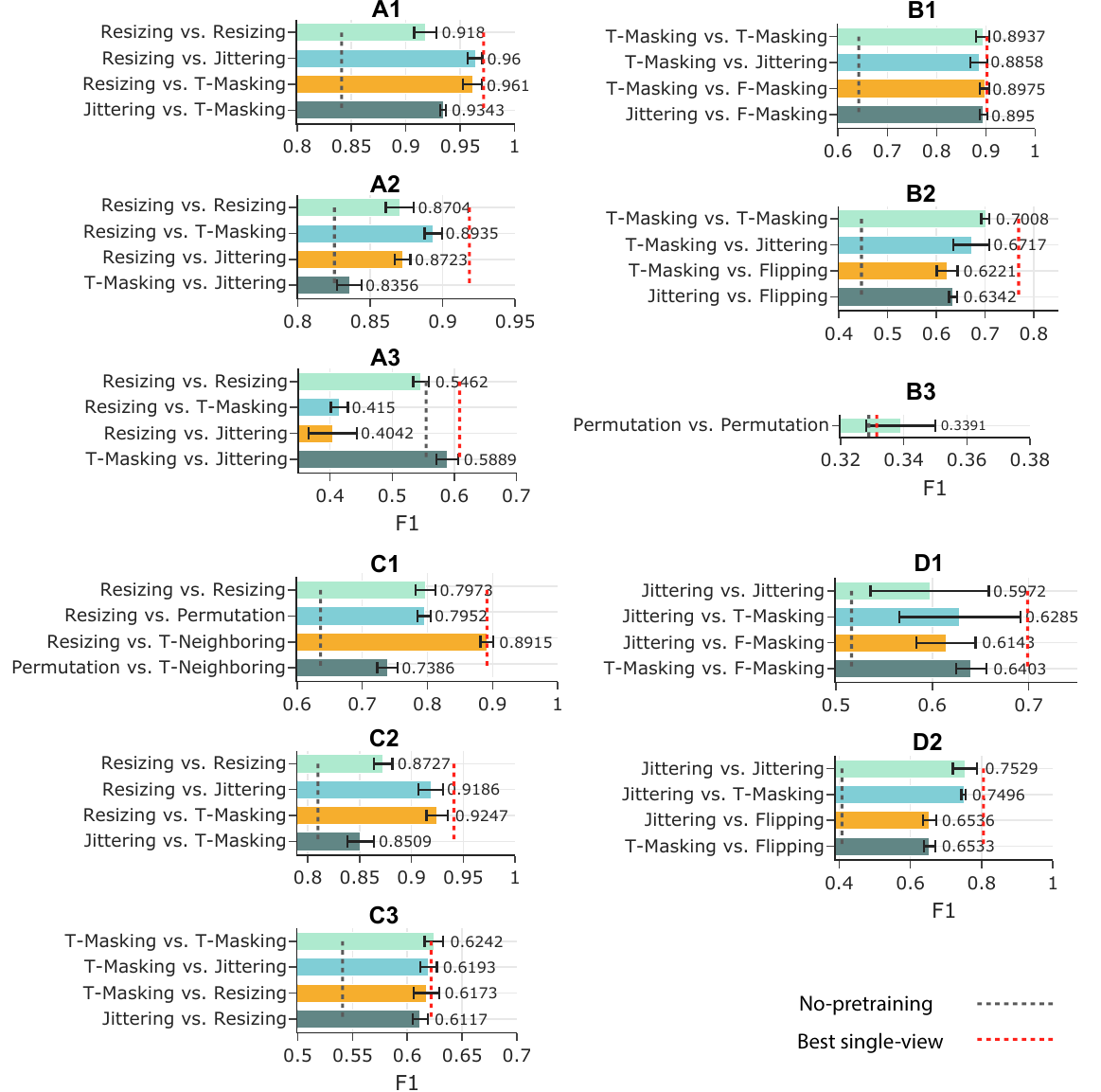}
    \caption{Performance summary of synthetic datasets on double-view augmentation benchmarking.}
    \label{fig:full_augset_bar}
\end{figure*}


\begin{figure*}
    \centering
    \includegraphics[width=0.95\textwidth]{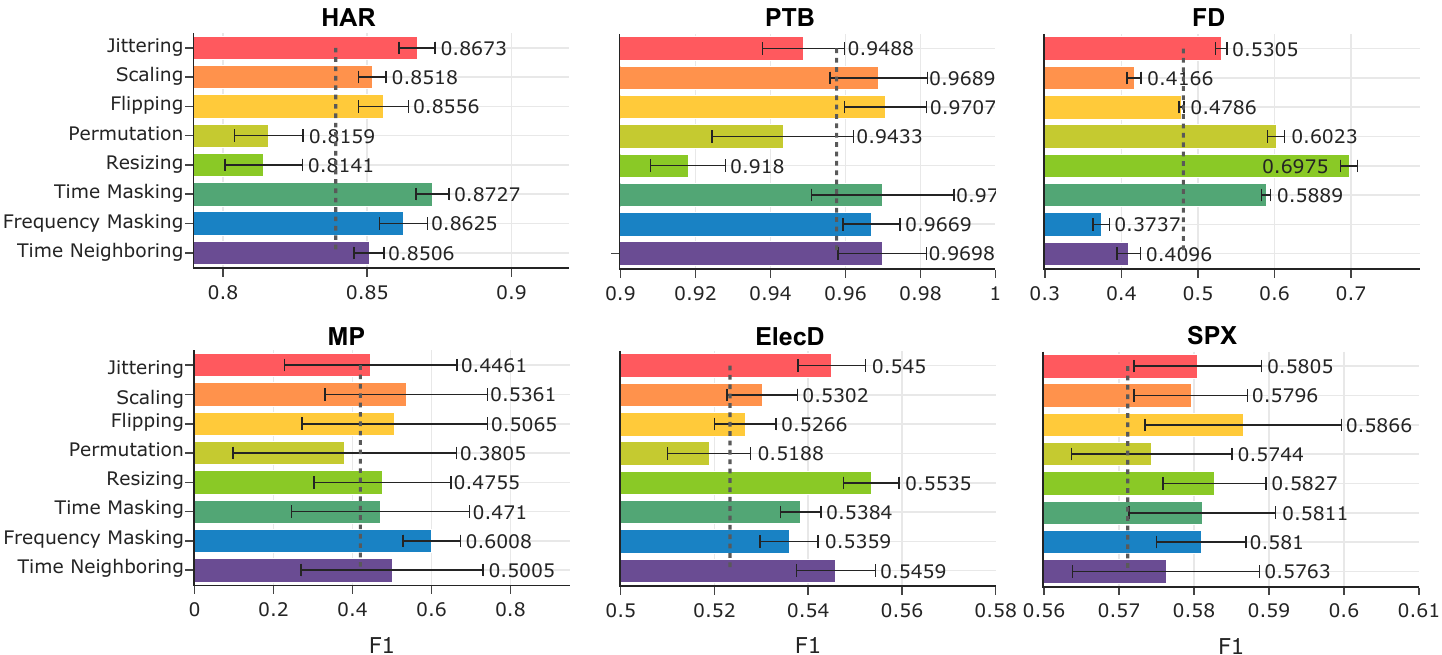}
    \caption{Performance summary of real-world datasets on original vs augmentation benchmarking.}
    \label{fig:performance_real_bar}
\end{figure*}



\begin{table*}[]
\centering
\caption{Full evaluation of augmentation recommendation.}
\label{tab:recall_at_k}
\resizebox{0.75\textwidth}{!}{%
\begin{tabular}{@{}lccccccc@{}}
\toprule
\multicolumn{2}{c}{\textbf{Recommendation method}} & \textbf{HAR} & \textbf{PTB} & \textbf{FD} & \textbf{MP} & \textbf{ElecD} & \textbf{SPX500} \\ \midrule
\multicolumn{1}{c}{\textbf{Random}} & Recall@1 & 0.113 & 0.107 & 0.108 & 0.101 & 0.108 & 0.101 \\
 & Recall@2 & 0.235 & 0.2175 & 0.222 & 0.216 & 0.218 & 0.2095 \\
 & Recall@3 & 0.335 & 0.3303 & 0.3357 & 0.3267 & 0.3387 & 0.3307 \\
 & Recall@4 & 0.445 & 0.4475 & 0.4435 & 0.4465 & 0.4465 & 0.4465 \\
 & Recall@5 & 0.5532 & 0.555 & 0.5566 & 0.5548 & 0.557 & 0.557 \\
 & Recall@6 & 0.6598 & 0.6645 & 0.6708 & 0.667 & 0.666 & 0.6653 \\
 & Recall@7 & 0.7756 & 0.7756 & 0.7776 & 0.7767 & 0.7767 & 0.7767 \\
 & Recall@8 & 0.8881 & 0.8881 & 0.888 & 0.8879 & 0.8879 & 0.8871 \\
 & Recall@9 & 1 & 1 & 1 & 1 & 1 & 1 \\ \midrule
\multicolumn{1}{c}{\textbf{Popularity-based}} & Recall@1 & 1 & 0 & 0 & 0 & 0 & 0 \\
 & Recall@2 & 1 & 0 & 0 & 0 & 0.5 & 0 \\
 & Recall@3 & 0.667 & 0.3333 & 0.667 & 0 & 0.667 & 0.667 \\
 & Recall@4 & 0.75 & 0.5 & 0.75 & 0.25 & 0.75 & 0.75 \\
 & Recall@5 & 0.6 & 0.6 & 0.8 & 0.6 & 0.8 & 0.8 \\
 & Recall@6 & 0.667 & 0.667 & 0.8333 & 0.667 & 0.667 & 0.667 \\
 & Recall@7 & 0.8571 & 0.8571 & 0.8571 & 0.8571 & 0.8571 & 0.8571 \\
 & Recall@8 & 0.875 & 0.875 & 0.875 & 1 & 1 & 0.875 \\
 & Recall@9 & 1 & 1 & 1 & 1 & 1 & 1 \\ \midrule
\multicolumn{1}{c}{\textbf{Trend-Season-Based}} & Recall@1 & 1 & 0 & 1 & 0 & 1 & 0 \\
 & Recall@2 & 1 & 0 & 1 & 0 & 1 & 0.5 \\
 & Recall@3 & 0.667 & 0.667 & 1 & 0 & 0.667 & 0.667 \\
 & Recall@4 & 0.5 & 0.5 & 0.75 & 0.25 & 0.75 & 0.5 \\
 & Recall@5 & 0.6 & 0.6 & 0.8 & 0.4 & 0.6 & 0.6 \\
 & Recall@6 & 0.667 & 0.667 & 0.667 & 0.667 & 0.667 & 0.667 \\
 & Recall@7 & 0.8571 & 0.8571 & 0.8571 & 0.7143 & 0.7143 & 0.8571 \\
 & Recall@8 & 0.875 & 1 & 1 & 0.875 & 0.875 & 0.875 \\
 & Recall@9 & 1 & 1 & 1 & 1 & 1 & 1 \\ \bottomrule
\end{tabular}%
}
\end{table*}

\end{document}